\documentclass[a4paper,fleqn]{cas-dc}

\usepackage{xcolor}
\usepackage[figuresright]{rotating}
\usepackage[numbers]{natbib}
\usepackage{graphicx}
\usepackage{subcaption}
\usepackage{caption}
\newcommand{\subsubsubsection}[1]{\paragraph{#1}\mbox{}\\}
\def\tsc#1{\csdef{#1}{\textsc{\lowercase{#1}}\xspace}}
\tsc{WGM}
\tsc{QE}
\begin{document}
\begin{sloppypar}
\let\WriteBookmarks\relax
\def\floatpagepagefraction{1}
\def\textpagefraction{.001}
\let\printorcid\relax

% Short title
\shorttitle{MOSMOS: Multi-organ segmentation facilitated by medical report supervision}    

% Short author
\shortauthors{W. Tian \MakeLowercase{\textit{et al.}}}  

% Main title of the paper
\title [mode = title]{MOSMOS: Multi-organ segmentation facilitated by medical report supervision}  

% Title footnote mark
% eg: \tnotemark[1]
% \tnotemark[<tnote number>] 

% Title footnote 1.
% eg: \tnotetext[1]{Title footnote text}
% \tnotetext[<tnote number>]{<tnote text>} 

% First author
%
% Options: Use if required
% eg: \author[1,3]{Author Name}[type=editor,
%       style=chinese,
%       auid=000,
%       bioid=1,
%       prefix=Sir,
%       orcid=0000-0000-0000-0000,
%       facebook=<facebook id>,
%       twitter=<twitter id>,
%       linkedin=<linkedin id>,
%       gplus=<gplus id>]

\author[1]{\textcolor{black}{Weiwei} Tian}
\ead{wwtian20@fudan.edu.cn}

\author[2]{\textcolor{black}{Xinyu} Huang}
\ead{xinyuhuang20@fudan.edu.cn}

\author[2,5]{\textcolor{black}{Junlin} Hou}
\ead{csejlhou@ust.hk}

\author[2]{\textcolor{black}{Caiyue} Ren}
\ead{rencaiyue@163.com}

\author[2]{\textcolor{black}{Longquan} Jiang}
\ead{lqjiang@fudan.edu.cn}
\cormark[1]
\cortext[1]{Corresponding author}

\author[1]{\textcolor{black}{Rui-Wei} Zhao}
\ead{rwzhao@fudan.edu.cn}

\author[4]{\textcolor{black}{Gang} Jin}
\ead{jingang@smmu.edu.cn}

\author[2,3]{\textcolor{black}{Yuejie} Zhang}
\ead{yjzhang@fudan.edu.cn}

\author[1]{\textcolor{black}{Daoying} Geng}
\ead{gengdy@163.com}

\affiliation[1]{organization={Academy for Engineering and Technology, Fudan University},
            city={Shanghai},
            postcode={200433}, 
            country={China}}

\affiliation[2]{organization={School of Computer Science, Shanghai Key Laboratory of Intelligent Information Processing, Fudan University},
            city={Shanghai},
            postcode={200433},
            country={China}}

\affiliation[3]{organization={Shanghai Collaborative Innovation Center of Intelligent Visual Computing},
            country={China}}

\affiliation[4]{organization={Department of Hepatobiliary Pancreatic Surgery, Changhai Hospital, Second Military Medical University~(Naval Medical University)},
            city={Shanghai},
            postcode={200433},
            country={China}}

\affiliation[5]{organization={Department of Computer Science and Engineering, The Hong Kong University of Science and Technology},
            country={China}}

% Corresponding author indication
% \cormark[<corr mark no>]

% Footnote of the first author
% \fnmark[<footnote mark no>]

% Email id of the first author
% \ead{<email address>}

% URL of the first author
% \ead[url]{<URL>}

% Credit authorship
% eg: \credit{Conceptualization of this study, Methodology, Software}
% \credit{<Credit authorship details>}

% Address/affiliation
% \affiliation[<aff no>]{organization={},
%             addressline={}, 
%             city={},
% %          citysep={}, % Uncomment if no comma needed between city and postcode
%             postcode={}, 
%             state={},
%             country={}}

% \author[<aff no>]{<author name>}

% Footnote of the second author
% \fnmark[2]

% Email id of the second author
% \ead{}

% URL of the second author
% \ead[url]{}

% Credit authorship
% \credit{}

% Address/affiliation
% \affiliation[<aff no>]{organization={},
%             addressline={}, 
%             city={},
% %          citysep={}, % Uncomment if no comma needed between city and postcode
%             postcode={}, 
%             state={},
%             country={}}

% Corresponding author text
% \cortext[1]{Corresponding author}

% Footnote text
% \fntext[1]{}

% For a title note without a number/mark
%\nonumnote{}

% Here goes the abstract
\begin{abstract}
Owing to a large amount of multi-modal data in modern medical systems, such as medical images and reports, Medical Vision-Language Pre-training~(Med-VLP) has demonstrated incredible achievements in coarse-grained downstream tasks~(i.e., medical classification, retrieval, and visual question answering). However, the problem of transferring knowledge learned from Med-VLP to fine-grained multi-organ segmentation tasks has barely been investigated. Multi-organ segmentation is challenging mainly due to the lack of large-scale fully annotated datasets and the wide variation in the shape and size of the same organ between individuals with different diseases. In this paper, we propose a novel pre-training \& fine-tuning framework for Multi-Organ Segmentation by harnessing Medical repOrt Supervision~(MOSMOS). Specifically, we first introduce global contrastive learning to maximally align the medical image-report pairs in the pre-training stage. To remedy the granularity discrepancy, we further leverage multi-label recognition to implicitly learn the semantic correspondence between image pixels and organ tags. More importantly, our pre-trained models can be transferred to any segmentation model by introducing the pixel-tag attention maps. Different network settings, i.e., 2D U-Net and 3D UNETR, are utilized to validate the generalization. We have extensively evaluated our approach using different diseases and modalities on BTCV, AMOS, MMWHS, and BRATS datasets. Experimental results in various settings demonstrate the effectiveness of our framework. This framework can serve as the foundation to facilitate future research on automatic annotation tasks under the supervision of medical reports.
\vspace{-1.0em}
\end{abstract}

% Use if graphical abstract is present
%\begin{graphicalabstract}
%\includegraphics{}
%\end{graphicalabstract}

% Research highlights
% \begin{highlights}
% \item 
% \item 
% \item 
% \end{highlights}

% Keywords
% Each keyword is seperated by \sep

\maketitle

% Main text
\section{Introduction}
Assigning an organ tag to each pixel in a medical image, also known as multi-organ segmentation, is a crucial task in medical image analysis, as it contributes to various computer-aided diagnosis and treatment tasks, including volume measurement~\cite{sharbatdaran2022deep}, 3D reconstruction~\cite{wang2018fully}, and treatment planning~\cite{lin2021variance}. To achieve these clinical applications, it is necessary to segment multiple organs in medical images accurately and robustly. However, compared to one particular organ segmentation, manually annotating multiple organs by radiologists is not only time-consuming and laborious but also heavily dependent on their experience. Since automatic multi-organ segmentation is efficient, it becomes an essential issue to address the growing clinical needs~\cite{wang2019abdominal}.

\begin{figure}[!t]
\vspace{0.5em}
\centerline{\includegraphics[width=1.0\columnwidth, height=0.50\columnwidth]{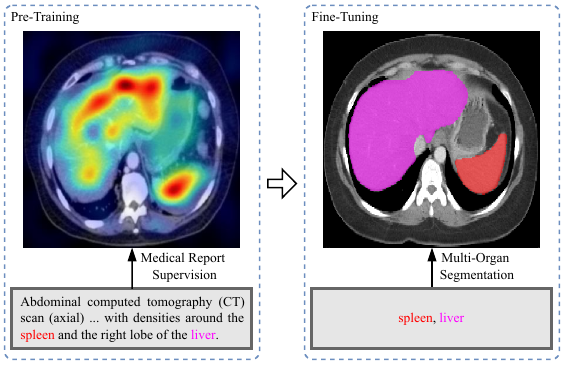}}
\caption{An example of our multi-organ segmentation result when using medical report supervision for pre-training. Left: The attention map locating the organ tags in the radiology image extracted from the corresponding medical report. Right: Our segmentation result for the corresponding organ tags.}
\label{intro}
\vspace{-1.5em}
\end{figure}

With the development of Fully Convolutional Networks~(FCN)~\cite{long2015fully} and Vision Transformers~(ViT)~\cite{dosovitskiy2020vit}, impressive segmentation performance has been achieved. However, existing works on multi-organ segmentation are usually based on the supervised learning paradigm~\cite{zhang2020block, zhang2021dodnet, cao2022swin, hatamizadeh2022unetr}, which is dramatically limited by high-quality and high-cost annotations. To tackle this issue, pre-training on large-scale datasets and then fine-tuning on smaller target datasets has become a widely adopted mode. For instance, Swin UNETR~\cite{tang2022self} leveraged self-supervised pre-training with tailored proxy tasks to alleviate the lack of annotations. Nevertheless, it only learns transferable visual representations from five Computer Tomography~(CT) datasets. It is unsuitable for segmentation tasks in other diseases or modalities, such as Magnetic Resonance Imaging~(MRI).

In summary, critical difficulties exist in two aspects with multi-organ segmentation: (i) There is a lack of large-scale fully annotated, multi-disease, or multi-modal datasets. (ii) The shape and size of the same organ vary significantly between patients with different diseases, making it difficult for the network to learn representative features. To address the first limitation, we argue that medical reports reflect radiologists' perceptions of multi-disease and multi-modal medical images, which can serve as weakly supervised information to help optimize the multi-organ segmentation network even with fewer annotations~(see Fig. \ref{intro}). Moreover, taking into consideration that radiologists prepare medical reports accompanied with radiology images as part of their daily routine, large-scale medical image-report pairs are easy to access without extra cost, in contrast to pixel-level fine-grained annotations. To address the second challenge, we simultaneously introduce global image-report aligning and local pixel-tag aligning to identify discriminative representations for the same organ with different diseases in the pre-training stage. Furthermore, we design pixel-tag attention maps to assist multi-organ segmentation tasks in the fine-tuning stage. 

Concretely, we propose a novel pre-training \& fine-tuning framework named \textbf{MOSMOS} for multi-organ segmentation based on medical report supervision. In the pre-training phase, image-report contrastive learning is used to align the global features of medical images and corresponding reports. In addition, we apply a more fine-grained pre-training task called multi-label recognition. It can locate image regions with the organ tags in the corresponding reports, which has the following advantages: (i) The tags are the organ classification labels extracted from the reports without additional manual annotations. (ii) The tags are encoded into query embeddings and then fed into a Transformer decoder~\cite{vaswani2017attention, liu2021query2label, zhang2023recognize} to perform multi-modal interaction, which guarantees the generalizability of the pre-trained model transferred to multi-disease, multi-modal, and multi-organ segmentation tasks. (iii) By implicitly optimizing the attention maps in the Transformer decoder, the organ tags can be associated with fine-grained and interpretable location information. They are capable of assisting multi-organ segmentation tasks to be better optimized since attention maps can also be regarded as segmentation results with low resolution. In the fine-tuning phase, we combine the segmentation loss and the pixel-tag aligning loss to supervise the training process.

Our pre-trained model can be fine-tuned on any downstream segmentation framework to boost performance. A series of comprehensive experiments have proved the effectiveness of our method. In the aspect of the downstream segmentation frameworks, we verify on the representative segmentation models (U-Net~\cite{ronneberger2015u} \& UNETR~\cite{hatamizadeh2022unetr}) with two mainstream visual backbones (ResNet~\cite{he2016deep} \& ViT~\cite{dosovitskiy2020vit}), respectively. As for the downstream segmentation datasets, we evaluate on four publicly available multi-disease and multi-organ datasets (BTCV~\cite{landman2015miccai} \& AMOS~\cite{ji2022amos} \&  MMWHS~\cite{zhuang2016multi} \& BRATS~\cite{simpson2019large}) with different modalities (CT \& MRI).

The main contributions of this work are summarized as follows:
\begin{itemize}

\item We establish MOSMOS, a novel pre-training \& fine-tuning framework to fully leverage the intrinsic medical report supervision within the paired images and reports to learn medical visual representation instead of purely exploiting radiology images. To the best of our knowledge, this is the first work that the medical vision-language pre-training is applied to downstream tasks of multi-organ segmentation. 

\item We design global image-report aligning and local pixel-tag aligning in the pre-training stage, which is more suitable for fine-grained segmentation tasks in the downstream.

\item We verify the effectiveness of the proposed method on the representative segmentation frameworks and four widely used multi-disease and multi-organ datasets of different modalities with 2D and 3D medical images. Our proposed MOSMOS significantly improves the multi-organ segmentation performance by a substantial margin.

\end{itemize}

\vspace{-1.0em}
\section{Related work}
Before introducing the proposed method, we mainly review previous works that inspired the design of our multi-organ segmentation scheme in this section. The two essential parts are (i) multi-organ segmentation; (ii) language supervision, in order to leverage cross-modal information to guide the multi-organ segmentation. 

\subsection{Multi-organ segmentation}
Many attempts have been made to implement multi-organ segmentation more efficiently. According to the backbone, these approaches can be divided into three categories. (i) FCN-based: To leverage the partially labeled datasets, the multi-head strategy~\cite{chen2019med3d, fang2020multi, shi2021marginal} was used for segmentation, which consists of a task-shared encoder and multiple decoders~(layers) with specific tasks, leading to poor scalability. To improve the flexibility, DoDNet~\cite{zhang2021dodnet} built a dynamic on-demand framework that introduced a dynamic segmentation head to the shared encoder-decoder structure. (ii) ViT-based: Swin-Unet~\cite{cao2022swin} first utilized hierarchical Swin Transformer~\cite{liu2021swin} with shifted window operation to capture global and long-term semantic information. (iii) FCN and ViT combined: Taking advantage of the locality of convolution and the globality of self-attention in Transformer, recent works~\cite{chen2021transunet, xie2023learning, hatamizadeh2022unetr, zhou2021nnformer} adopted the hybrid architecture. Based on U-Net~\cite{ronneberger2015u} architecture, TransUNet~\cite{chen2021transunet} and TransDoDNet~\cite{xie2023learning} introduced Transformer as a bottleneck feature extractor for modeling long-range organ-wise dependencies, which is conducive to multi-organ segmentation. UNETR~\cite{hatamizadeh2022unetr} used Transformer as the encoder and delivered the encoded representations to the FCN-based decoder by skip connections. NnFormer~\cite{zhou2021nnformer} applied interleaved convolutional layers and Transformer blocks to play both advantages sufficiently. However, the performance of these supervised learning methods learning from scratch is limited by the quantity and quality of annotations, or that transferring pre-trained weights from ImageNet~\cite{deng2009imagenet} is suboptimal due to the drastic difference between natural and medical images. Performance improvements have been achieved through supervised learning methods that transferred pre-trained weights from large-scale, partially labeled medical datasets. Nonetheless, these methods also need intensive labor and expertise costs.

Recent advances in self-supervised pre-training~\cite{chen2019self, taleb20203d, zhu2020rubik, tang2022self, xie2024refs} provided the promise of leveraging unlabeled medical images. Specifically, Swin UNETR~\cite{tang2022self} first designed three tailored proxy tasks, that is, masked volume inpainting, rotation prediction, and contrastive learning, to pre-train the Swin Transformer encoder. The pre-trained encoder was transferred to downstream segmentation tasks and achieved observable improvements. Despite its success, a gap exists between the upstream self-supervised task and the downstream segmentation tasks. Consequently, ReFs~\cite{xie2024refs} proposed an extra supervised reference task as a bridge to minimize the gap. Unlike these approaches, our pre-training framework introduces the cross-modal supervisory information in paired medical images and reports at no extra cost to facilitate multi-disease, multi-modal, and multi-organ segmentation tasks. Meanwhile, we employ multi-label recognition to align image pixels with organ tags automatically extracted from medical reports, bridging the gap between upstream and downstream tasks.

\subsection{Language supervision}
Towards the goal of utilizing unlabeled images more efficiently, several follow-ups~\cite{jia2021scaling, xu2021simple, rao2022denseclip, huang2022idea, xu2022groupvit, huang2023tag2text, huang2023open} based on Contrastive Language-Image Pre-training~(CLIP)~\cite{radford2021learning} have achieved promising results in learning visual representation with language supervision using plenty of image-text pairs in the general domain. Inspired by these pioneering works, \cite{zhang2022contrastive, eslami2021does, zhou2022generalized, seibold2022breaking, chen2022align, wang2022medclip, liao2021multimodal} applied modified CLIP to medical classification, retrieval, and visual question answering tasks. For more fine-grained dense prediction tasks, LViT~\cite{li2023lvit} introduced medical text annotations to lead the generation of pseudo labels in semi-supervised learning. In addition to using global contrastive learning to align medical images and reports, GLoRIA~\cite{huang2021gloria} and LoVT~\cite{muller2022joint} proposed utilizing local contrastive learning to align image sub-regions and words or sentences in the paired reports, and BioViL~\cite{boecking2022making} adopted masked language modeling to leverage text semantics sufficiently. Furthermore, MGCA~\cite{wang2022multi} explored the abundant semantic correspondences between radiology images and reports with multiple granularities: disease-level, instance-level, and token-level. Despite achieving exceptional performance, these segmentation or detection approaches that utilize language supervision are confined to the localization of pulmonary lesions or cell nuclei in 2D images. Augmenting the above-mentioned medical segmentation methods, we extend to broader multi-organ segmentation scenarios of different modalities with 2D and 3D medical images by introducing global image-report aligning and local pixel-tag aligning using multi-label recognition in the pre-training stage.

\begin{figure}[!t]
\centerline{\includegraphics[width=0.6\columnwidth,height=0.6\columnwidth]{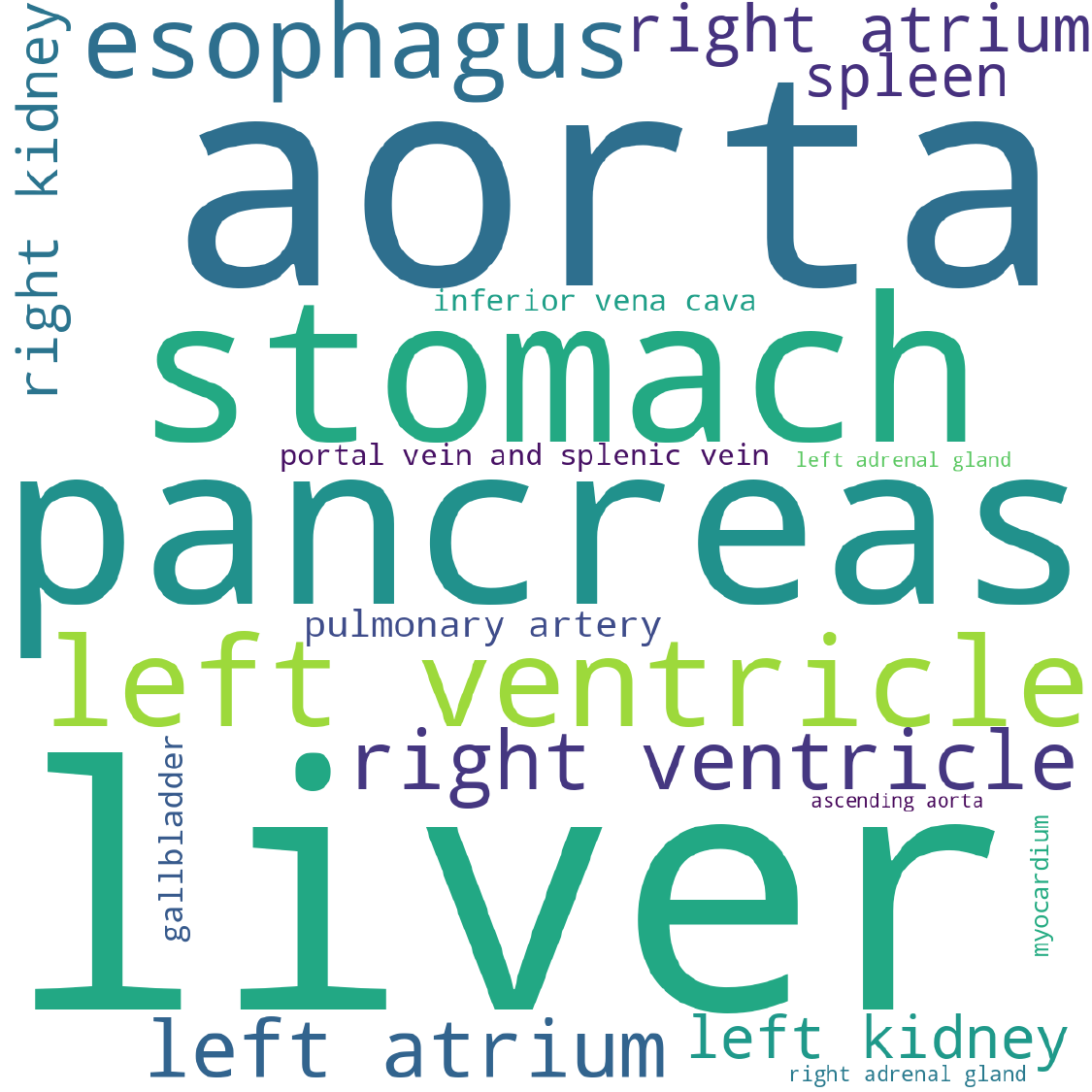}}
\caption{Illustration of the 20 organ categories in the tag list. The tag size is proportional to the tag frequency in the training set of the ROCO dataset.}
\label{tags}
\vspace{-1.5em}
\end{figure}

\section{Material and methods}

In this section, we first present the datasets and the overview of our MOSMOS framework. Next, we introduce the pre-training method of MOSMOS, including global image-report aligning and local pixel-tag aligning. Then we illustrate the fine-tuning approach, which utilizes weakly supervised positioning to facilitate multi-organ segmentation.

\subsection{Datasets}

\subsubsection{Dataset for pre-training\label{tag_list}}
The Radiology Objects in COntext~(ROCO) dataset~\cite{pelka2018radiology} contains over 81,000 2D radiology images, split into 73,594 and 8,176 images for training and validation sets, respectively. ROCO does not concentrate on a specific disease or anatomical structure but addresses multi-modal radiology images, including Angiography, CT, Fluoroscopy, MRI, Mammography, Positron Emission Tomography~(PET), PET-CT, Ultrasound, and X-Ray.

All images in ROCO have corresponding medical reports and a set of organ tags obtained from the reports. Each report describes the visual element in its semantic context. To acquire the organ tags for radiology images, we first define $K=20$ common organ categories, including abbreviations and synonyms. Then the list is double-checked by radiologists. After substituting abbreviations and synonyms with the unified forms, we extract the organ tags from the medical reports by matching. An aggregation step is further executed to merge the multiple mentioned tags in a report. A detailed overview of this tag list is shown in Fig. \ref{tags}.

\begin{figure*}[thbp]
\begin{center}
\includegraphics[width=\textwidth,height=0.74\textwidth]{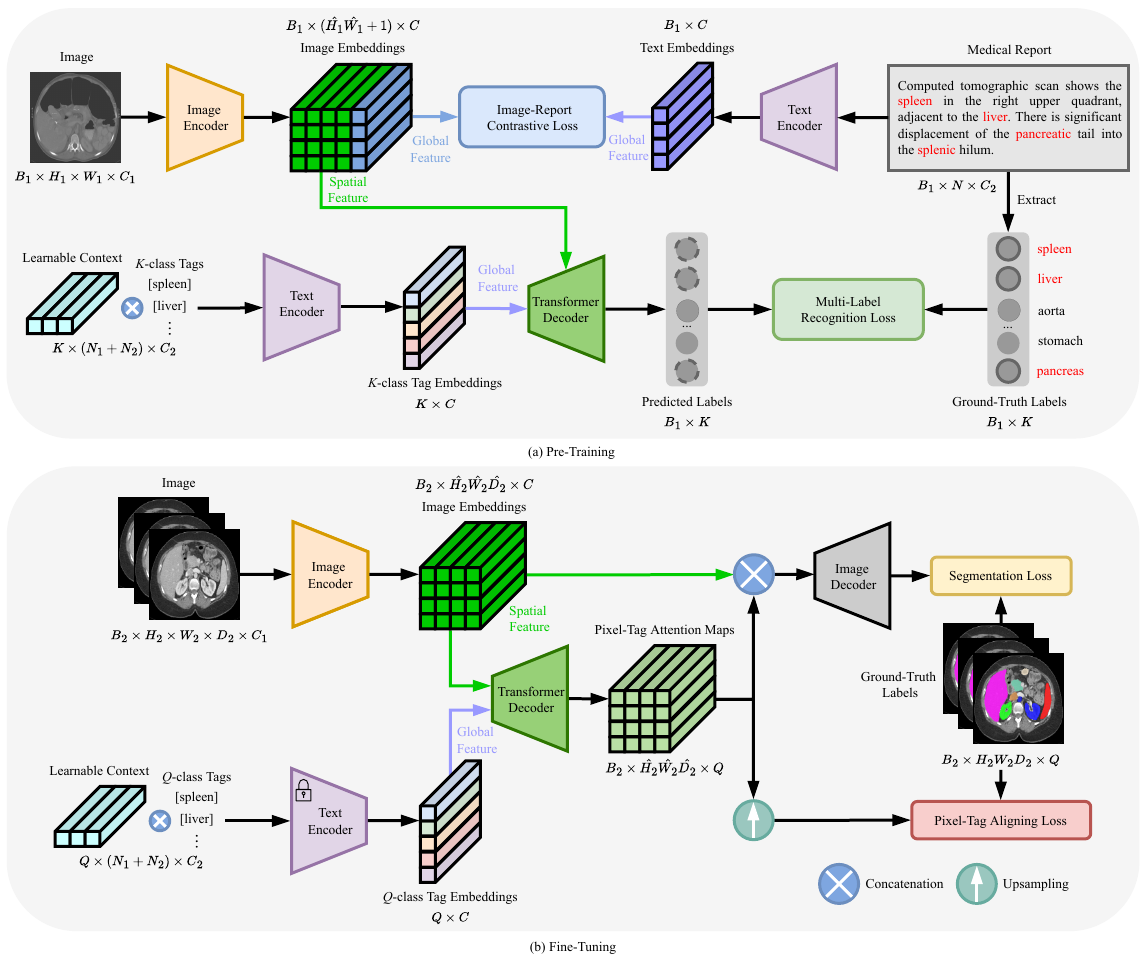}
\end{center}
\vspace{-1.0em}
\caption{Illustration of our proposed MOSMOS framework in both pre-training and fine-tuning stages. (a) In the pre-training stage, MOSMOS applies image-report contrastive learning to align the global features of radiology images with those of corresponding medical reports. To further learn fine-grained visual representation from medical report supervision, the visual spatial features and the embeddings of the constructed $K$-class tags are sent to the Transformer decoder for multi-label recognition. Note that the ground-truth tags are extracted from the medical reports with no manual annotation. Note: $B_{1}$: batch size, $H_{1}$: height of the image, $W_{1}$: width of the image, $C_{1}$: dimension of the image, $\hat{H}_{1}$: height of the image embedding, $\hat{W}_{1}$: width of the image embedding, $C$: dimension of the image embedding, text embedding, and tag embedding, $K$: number of the organ tags, $N$: token length of the medical report, $N_{1}$: token length of the learnable textual context, $N_{2}$: token length of the tag, $C_{2}$: dimension of the medical report and tag. (b) In the fine-tuning stage, the pixel-tag attention maps calculated by the Transformer decoder are fed into the image decoder. The segmentation loss and the pixel-tag aligning loss are combined to supervise the training process. Note that the learnable textual context is shared across all tags and is continuously updated in both stages. Note: $B_{2}$: batch size, $H_{2}$: height of the image, $W_{2}$: width of the image, $D_{2}$: depth of the image, $\hat{H}_{2}$: height of the image embedding, $\hat{W}_{2}$: width of the image embedding, $\hat{D}_{2}$: depth of the image embedding, $Q$: number of the organ tags. For 2D images, $D_{2}$ and $\hat{D}_{2}$ are omitted.}
\label{network}
\vspace{-1.5em}
\end{figure*}

\subsubsection{Datasets for fine-tuning}
We extensively evaluate our multi-organ segmentation approach on two modalities of datasets from different human body regions, that is, BTCV~\cite{landman2015miccai} for abdominal multi-organ segmentation using CT, AMOS~\cite{ji2022amos} for abdominal multi-organ segmentation using CT and MRI, MMWHS~\cite{zhuang2016multi} for cardiac substructure segmentation using MRI, and BRATS~\cite{simpson2019large} for brain tumor segmentation using MRI. These datasets adopted for fine-tuning do not need to provide medical reports but human-assisted annotations. Following the split ratios of ~\cite{dou2020unpaired}, the percentages for training, validation, and test sets on BTCV, MMWHS, and BRATS are 70$\%$, 10$\%$, and 20$\%$, respectively. The AMOS dataset is divided into training and validation sets at a ratio of 2:1.

\begin{itemize}

\item \textbf{BTCV} provides annotations of $Q=13$ abdominal organs~(that is, spleen, right kidney, left kidney, gallbladder, esophagus, liver, stomach, aorta, inferior vena cava, portal vein and splenic vein, pancreas, right adrenal gland, and left adrenal gland). There are 30 abdomen CT scans from colorectal cancer or ventral hernia patients acquired during the portal venous contrast phase. All images are manually annotated and further verified by experienced radiologists from Vanderbilt University Medical Center.

\item \textbf{AMOS} consists of 300 CT and 60 MRI scans, collected from multi-center, multi-vendor, multi-phase, multi-disease patients. It provides voxel-level annotations for $Q=15$ abdominal organs, namely spleen, right kidney, left kidney, gallbladder, esophagus, liver, stomach, aorta, inferior vena cava, pancreas, right adrenal gland, left adrenal gland, duodenum, bladder, and prostate or uterus. Notably, the duodenum, bladder, and prostate or uterus are considered open-set organ categories, for which tags are not extracted during the pre-training stage. Additionally, MRI scans lack annotations for the bladder and prostate or uterus organs.

\item \textbf{MMWHS} is a dataset for whole heart segmentation of $Q=7$ cardiac substructures~(that is, myocardium, left atrium, left ventricle, right atrium, right ventricle, ascending aorta, pulmonary artery), containing 20 cardiac MRI images from patients with cardiovascular diseases. These data are obtained using 3D balanced steady-state free precession~(b-SSFP) sequences.

\item \textbf{BRATS} is specifically designed for brain tumor segmentation, which comprises 484 multi-modal MRI scans~(including FLAIR, T1w, T1gd, T2w modalities) from patients diagnosed with gliomas. All $Q=3$ segmentation targets~(that is, tumor core, whole tumor, enhancing tumor) are categorized as open-set.

\end{itemize}

\subsection{MOSMOS}
As shown in Fig. \ref{network}, MOSMOS is a two-stage framework for multi-organ segmentation based on medical report supervision. Given a batch of image-report pairs in the first pre-training stage, we first split the visual representations into global and spatial features through the image encoder, and the textual representations into report-level and tag-level features through the shared text encoder. Then we perform two tasks: global image-report aligning and local pixel-tag aligning. The first task adopts the contrastive learning strategy, which reinforces the matching degree between the visual global representations of the radiology images and the textual global representations of the corresponding reports. The second task leverages multi-label recognition to align the image regions and the organ tags in the original medical reports. For this purpose, we apply a Transformer decoder based on cross-attention to fully leverage visual spatial features to recognize tags located in the images. In the second fine-tuning stage, the pixel-tag attention maps generated by the Transformer decoder are concatenated with the visual spatial features and then fed into the image decoder concurrently. Besides the segmentation loss, the pixel-tag aligning loss is also applied to supervise the training process. 

\subsubsection{Pre-training}

\subsubsubsection{Global image-report aligning\label{irc}}
In the routine clinical workflow, medical reports paired with radiology images are generated naturally by experienced radiologists. Assume each image-report pair is unique. We utilize global image-report contrastive learning to align image-report representations. For a mini-batch of $B_{1}$ image-report pairs~$(I, R)$ sampled from training dataset, we use $(I_{i}, R_{i})$ to represent the $i$-th pair. We embed the 2D image $I_{i} \in \mathbb{R}^{H_{1} \times W_{1} \times C_{1}}$ with resolution $(H_{1}, W_{1})$ and $C_{1}$ input dimension via an image encoder $e^{I}$ and a linear projection layer $p^{I}$ into a global feature $f^{G}_{i} \in \mathbb{R}^{C}$ and a spatial feature $f^{S}_{i} \in \mathbb{R}^{\hat{H}_{1}\hat{W}_{1} \times C}$, where $(\hat{H}_{1}, \hat{W}_{1})$ and $C$ denote the resolution and dimension of the feature map, respectively:

\vspace{-1.0em}
\begin{equation}
f_{i}^{G}, f_{i}^{S}=p^{I}\left(e^{I}\left(I_{i}\right)\right).
\end{equation}

Following the similar processing pipeline, $R_{i} \in \mathbb{R}^{N \times C_{2}}$ with token length $N$ and $C_{2}$ dimension is converted into a $C$-dimension global representation $f^{R}_{i}$ by a text encoder $e^{T}$ and a linear projection function $p^{T}$:

\vspace{-1.0em}
\begin{equation}
f_{i}^{R}=p^{T}\left(e^{T}\left(R_{i}\right)\right).
\end{equation}

Note that our model is agnostic to the specific option of image and text encoders. Following previous work~\cite{radford2021learning}, we apply ResNet~\cite{he2016deep} and ViT~\cite{dosovitskiy2020vit} as the image encoders $e^{I}$. The main difference between them is that ResNet performs a global attention pooling on the spatial feature $f^{S}_{i}$ to obtain the global feature $f^{G}_{i}$, while $f^{G}_{i}$ of ViT is the corresponding output of $[class]$ token. As for the text encoder $e^{T}$, we follow the encoder part of the Transformer~\cite{vaswani2017attention} architecture. The projectors $p^{I}$ and $p^{T}$ map the representations of images and medical reports into the same space of $C$ dimension so that contrastive learning can be applied. Based on the bidirectional image-to-report and report-to-image InfoNCE losses~\cite{oord2018representation}, the global image-report contrastive loss for each training mini-batch can be formulated as:

\vspace{-1.0em}
\begin{equation}
\mathcal{L}_{\text{i2r}} = -\log \frac{e^{\cos \left(f_{i}^{G}, f_{i}^{R}\right) / \tau}}{\sum_{j=1}^{B_{1}} e^{\cos \left(f_{i}^{G}, f_{j}^{R}\right) / \tau}},
\end{equation}

\vspace{-1.0em}
\begin{equation}
\mathcal{L}_{\text{r2i}} = -\log \frac{e^{\cos \left(f_{i}^{R}, f_{i}^{G}\right) / \tau}}{\sum_{j=1}^{B_{1}} e^{\cos \left(f_{i}^{R}, f_{j}^{G}\right) / \tau}},
\end{equation}

\vspace{-1.0em}
\begin{equation}
\mathcal{L}_{\text{irc}} = \frac{1}{2 B_{1}} \sum_{i=1}^{B_{1}}\left(\mathcal{L}_{\text{i2r}}+\mathcal{L}_{\text{r2i}}\right),
\end{equation}
where $\cos \left(\cdot,\cdot\right)$ denotes the cosine similarity, $\cos\left(f_{i}^{G}, f_{i}^{R}\right) = \left(f_{i}^{G}\right)^{\top} f_{i}^{R} /\|f_{i}^{G}\|\|f_{i}^{R}\|$, $\top$ represents the transpose operation, $\|\cdot\|$ denotes the L2 normalization, and $\tau$ is the learnable temperature parameter and initializes to 0.07 following~\cite{radford2021learning}.

Furthermore, compared with natural image-text pairs~\cite{radford2021learning, jia2021scaling}, the publicly available medical multi-modal datasets~\cite{pelka2018radiology, demner2016preparing, johnson2019mimic} are relatively small to train a generalizable model. Thus we employ CLIP model parameters for initialization. A multitude of medical image-report pairs are subsequently employed for the purpose of fine-tuning CLIP within the medical domain.

\subsubsubsection{Local pixel-tag aligning}
In the pre-training stage, we expect to gain more language supervision to learn medical visual representations. The global image-report contrastive learning, however, mainly considers coarse-grained representations of both images and medical reports, while downstream tasks of multi-organ segmentation are pixel-level. To narrow the substantial gap between these stages, we introduce multi-label recognition to implicitly align the image pixels and the organ tags to obtain more fine-grained information.

Multi-label recognition predicts whether each organ tag exists in the radiology images. Unlike the original Query2Label~\cite{liu2021query2label} that directly used learnable label embeddings as the input queries, we introduce $K$-class tags as the input, which can be transferred to downstream segmentation tasks based on medical report supervision better. The details of constructing the tag list can be found in Sec~\ref{tag_list}. Motivated by CoOp~\cite{zhou2022learning}, we apply the learnable textual context to mitigate the domain gap between tags and medical reports. Then the input of the shared text encoder $e^{T}$ becomes:

\vspace{-1.0em}
\begin{equation}
T_{k} = \left \langle p_{k}, t_{k}\right \rangle, \quad 1 \leq k \leq K,
\end{equation}
where $p_{k} \in \mathbb{R}^{N_{1} \times C_{2}}$ is the learnable textual context, shared in $K$-class tags. $t_{k} \in \mathbb{R}^{N_{2} \times C_{2}}$ is the embedding of $k$-th organ tag, $\left \langle \cdot,\cdot\right \rangle$ denotes the concatenation, and $N_{1}$ and $N_{2}$ are the token lengths of the learnable textual context and the tag, respectively. Similar to the procedure of medical reports, we get the global representation $f^{T}_{k} \in \mathbb{R}^{C}$ of the tag:

\vspace{-1.0em}
\begin{equation}
f_{k}^{T}=p^{T}\left(e^{T}\left(T_{k}\right)\right).
\end{equation}

On the basis of the spatial feature $f^{S}_{i}$ of the input radiology image obtained in Sec~\ref{irc}, we treat $f^{T} \in \mathbb{R}^{K \times C}$ as queries and leverage the cross-attention mechanism in Transformer decoder~\cite{vaswani2017attention} to progressively integrate category-related contextualized information from the input image into the query embeddings:

\vspace{-1.0em}
\begin{equation}
f_{i}^{TS}=\text{TransDecoder}\left(f^{T},f_{i}^{S},f_{i}^{S}\right),
\end{equation}
where $f_{i}^{TS} \in \mathbb{R}^{K \times C}$ are the updated queries. To perform multi-label recognition, we regard predicting each label as a binary classification task and map the feature $f_{i,k}^{TS} \in \mathbb{R}^{C}$ for $k$-th category of $i$-th sample into a logit value applying a linear projection layer $p^{TS}$ followed by a sigmoid function:

\vspace{-1.0em}
\begin{equation}
y_{i,k}=\text{Sigmoid}\left(p^{TS}\left(f_{i,k}^{TS}\right)\right),
\end{equation}
where $y_{i,k} \in \left[0,1\right]$ is the predicted probability for $k$-th category of $i$-th sample. We denote the ground-truth labels of input image $I_{i}$ as $x_{i} = \left[x_{i,1},\cdots,x_{i,K}\right]$ where $x_{i,k} \in \left\{0,1\right\}$ is a discrete binary label. $x_{i,k} = 1$ if the $k$-th organ tag presents in the corresponding medical report $R_{i}$, otherwise $x_{i,k} = 0$. Medical reports usually only describe organs that appear abnormal on radiology images, so there may be plenty of false negative labels. To address this issue, we adopt a simple and effective loss, that is, weak assume negative loss~\cite{cole2021multi}, which introduces a weight parameter $\gamma \in \left[0,1\right]$ based on binary cross-entropy loss to reduce the effect of false negatives. For a training mini-batch, the multi-label recognition loss is defined as:

\vspace{-1.0em}
\begin{equation}
\mathcal{L}_{\text{mlr}}=-\frac{1}{B_{1}K} \sum_{i=1}^{B_{1}} \sum_{k=1}^{K}\left\{\begin{matrix}
\log\left(y_{i,k}\right), & x_{i,k}=1,\\
\gamma \log \left(1-y_{i,k}\right), & x_{i,k}=0, 
\end{matrix}\right.
\end{equation}
where $\gamma = 1/\left(K-1\right) $ ensures that the approximate single positive label has the same impact on the loss as the $K-1$ assumed negatives.

Formally, we minimize the total loss function of pre-training tasks of MOSMOS as:

\vspace{-1.0em}
\begin{equation}
\mathcal{L}_{\text{total\_up}}=\mathcal{L}_{\text{irc}} + \mathcal{L}_{\text{mlr}}.
\end{equation}

\subsubsection{Fine-tuning}
Since MOSMOS learns visual representations from medical report supervision in the pre-training stage, we would like to explore the effect of transferring the pre-trained model to multi-organ segmentation tasks. Note that our framework is model-agnostic. For our investigation, we consider two main medical segmentation methods, 2D U-Net~\cite{ronneberger2015u} and 3D UNETR~\cite{hatamizadeh2022unetr}, that adopt ResNet~\cite{he2016deep} and ViT~\cite{dosovitskiy2020vit} as their image encoders, respectively. To evaluate the contribution of MOSMOS, we substitute the image encoders with the pre-trained ones and introduce the pre-trained language supervision for multi-label recognition without the classifier.

\subsubsubsection{Weakly supervised positioning}
Thanks to the cross-attention mechanism in Transformer decoder~\cite{vaswani2017attention}, the generated pixel-tag attention maps can provide weakly supervised information. Specifically, the attention maps incorporate language supervision into medical visual representations and roughly locate the spatial distribution of the organ tags in the medical images. Take a $B_{2}$ mini-batch of 3D radiology images $\bar{I} \in \mathbb{R}^{B_{2} \times H_{2} \times W_{2} \times D_{2} \times C_{1}}$ and corresponding $Q$-class organ tag embeddings $\bar{T} \in \mathbb{R}^{Q \times \left(N_{1}+N_{2}\right) \times C_{2}}$, for example. We obtain the spatial features of images $\bar{f}^{S} \in \mathbb{R}^{B_{2} \times \hat{H}_{2}\hat{W}_{2}\hat{D}_{2} \times C}$ and the global features of tags $\bar{f}^{T} \in \mathbb{R}^{Q \times C}$ using the pre-trained image and text encoders followed by corresponding projectors, respectively, where $H_{2} \times W_{2} \times D_{2}$ and $\hat{H}_{2}\hat{W}_{2}\hat{D}_{2}$ represent the height, width, and depth of the input images and feature maps, respectively. Regarding $\bar{f}^{T}$ as queries and $\bar{f}^{S}$ as keys and values, we pass these features to the Transformer decoder and gain the pixel-tag attention maps $\bar{f}^{M} \in \mathbb{R}^{B_{2} \times \hat{H}_{2}\hat{W}_{2}\hat{D}_{2} \times Q}$:

\vspace{-1.0em}
\begin{equation}
\bar{f}^{M}=\text{TransDecoder}\left(\bar{f}^{T},\bar{f}^{S},\bar{f}^{S}\right).
\end{equation}

The attention maps represent the degree of pixel-tag aligning, which play a significant role in our framework. Firstly, the attention maps can be concatenated with the visual spatial features to integrate medical language prior to guide the segmentation, that is, $\bar{f}^{SM}=\left \langle \bar{f}^{S},\bar{f}^{M}\right \rangle \in \mathbb{R}^{B_{2} \times \hat{H}_{2}\hat{W}_{2}\hat{D}_{2} \times \left(C+Q\right)}$, and then fed into the image decoder. We obtain the predicted output $Y_{\text{seg}} \in \mathbb{R}^{B_{2} \times H_{2}W_{2}D_{2} \times Q}$. Secondly, we can regard the attention maps as the segmentation results with lower resolution, and thus upsample them to the original resolution by linear interpolation LI to calculate a pixel-tag aligning loss:

\vspace{-1.0em}
\begin{equation}
Y_{\text{pta}}=\text{LI}\left(\bar{f}^{M}/\varepsilon\right),
\end{equation}
where $Y_{\text{pta}} \in \mathbb{R}^{B_{2} \times H_{2}W_{2}D_{2} \times Q}$ is the pixel-tag aligning output, and $\varepsilon$ denotes a learnable temperature coefficient and initializes to $0.07$ following~\cite{he2020momentum}.

\subsubsubsection{Multi-organ segmentation}
In addition to the segmentation loss $\mathcal{L}_{\text{seg}}$, we propose a pixel-tag aligning loss $\mathcal{L}_{\text{pta}}$ to make better use of the pixel-tag attention maps and help dense segmentation tasks converge faster. Both losses are a combination of cross-entropy loss and dice loss~\cite{milletari2016v}:

\vspace{-1.0em}
\begin{equation}
\begin{aligned}
\mathcal{L}\left(X,Y\right)=\frac{1}{B_{2}}\sum_{b=1}^{B_{2}}\left(1-\frac{1}{V}\sum_{v=1}^{V}\sum_{q=1}^{Q}X_{b,v,q}\log Y_{b,v,q} \right.
\\
\left. -\frac{2}{Q}\sum_{q=1}^{Q}\frac{\textstyle\sum_{v=1}^{V}X_{b,v,q}Y_{b,v,q}}{\textstyle\sum_{v=1}^{V}X_{b,v,q}^{2}+\textstyle\sum_{v=1}^{V}Y_{b,v,q}^{2}}\right),
\end{aligned}
\end{equation}

\vspace{-1.0em}
\begin{equation}
\mathcal{L}_{\text{seg}}=\mathcal{L}\left(X,Y_{\text{seg}}\right),
\end{equation}

\vspace{-1.0em}
\begin{equation}
\mathcal{L}_{\text{pta}}=\mathcal{L}\left(X,Y_{\text{pta}}\right),
\end{equation}
where $X \in \mathbb{R}^{B_{2} \times H_{2}W_{2}D_{2} \times Q}$ and $Y \in \mathbb{R}^{B_{2} \times H_{2}W_{2}D_{2} \times Q}$ denote the one-hot encoded ground truth and the predicted output, respectively, and $V$ is the number of pixels.

The final loss function is a linear combination of the above two parts:

\vspace{-1.0em}
\begin{equation}
\mathcal{L}_{\text{total\_down}}=\mathcal{L}_{\text{seg}} + \lambda\mathcal{L}_{\text{pta}},
\label{eq_total}
\end{equation}
where $\lambda$ is the hyper-parameter to balance the two-part losses.

\vspace{-0.5em}
\section{Results}
In this section, we present the experimental details and analyze the results to demonstrate the flexibility and generalization of our proposed multi-organ segmentation algorithm that is facilitated by medical report supervision.

\subsection{Implementation details}
We implement MOSMOS in PyTorch on a single NVIDIA V100 GPU. Two segmentation baselines are considered, that is, U-Net~\cite{ronneberger2015u} with ResNet-50~\cite{he2016deep} and UNETR~\cite{hatamizadeh2022unetr} with ViT-B/16~\cite{dosovitskiy2020vit} visual backbones. The textual backbone is the same text encoder as in the CLIP~\cite{rao2022denseclip}. For the sake of a comprehensive analysis, we compare our method with the following seven methods. To ensure a fair comparison, we implement the other methods using the same backbone and hyper-parameter settings as those applied in MOSMOS. The detailed hyper-parameters are listed in Table \ref{tab0}.

\begin{itemize}

\item \textbf{Random Init.}: The visual backbone of the baseline is initialized using default random initialization.

\item \textbf{ImageNet~\cite{deng2009imagenet} Init.}: The visual backbone of the baseline is initialized with weights pre-trained on ImageNet.

\item \textbf{Inpainting+Contrast+Rotation~\cite{tang2022self}}: The visual backbone of the baseline is pre-trained through the utilization of three self-supervised proxy tasks on ROCO images, specifically, mask volume inpainting, contrastive learning, and rotation prediction.

\item \textbf{CLIP~\cite{radford2021learning}}: The visual backbone of the baseline is initialized with weights pre-trained on CLIP. 

\item \textbf{CLIP+DenseCLIP~\cite{rao2022denseclip}}: The visual and textual backbones of the DenseCLIP are initialized with weights pre-trained on CLIP.

\item \textbf{PubMedCLIP~\cite{eslami2021does}}: The visual backbone of the baseline is initialized with weights pre-trained on PubMedCLIP.

\item \textbf{PubMedCLIP+DenseCLIP~\cite{rao2022denseclip}}: The visual and textual backbones of the DenseCLIP are initialized with weights pre-trained on PubMedCLIP.

\end{itemize}

\begin{table}[ht]
\caption{Hyper-parameters applied in our MOSMOS.}
\label{tab0}
\renewcommand\arraystretch{1.4}
\setlength{\tabcolsep}{0.25pt}
\begin{tabular}{ccc}
\hline
Symbol        & Hyper-Parameter                                             & Value \\ \hline
$\tau$        & the temperature parameter of InfoNCE loss         & 0.07  \\
$\varepsilon$ & the temperature parameter of linear interpolation & 0.07  \\
$\lambda$     & the weight of pixel-tag aligning loss                       & 0.8   \\ \hline
\end{tabular}
\vspace{-1.5em}
\end{table}

\begin{table*}[htbp]
\caption{Quantitative comparisons of segmentation performance using Dice~($\%$) metric on BTCV test set. The best results are bolded. The results of our approach are marked with a gray background color. We calculate the p-value between the average performance of our MOSMOS and PubMedCLIP+DenseCLIP~\cite{rao2022denseclip} in Dice metric. Note: Spl: spleen, RKid: right kidney, LKid: left kidney, Gal: gallbladder, Eso: esophagus, Liv: liver, Sto: stomach, Aor: aorta, IVC: inferior vena cava, Vein: portal vein and splenic vein, Pan: pancreas, RAG: right adrenal gland, LAG: left adrenal gland.}
\vspace{-1.0em}
\label{tab1}
\begin{center}
\renewcommand\arraystretch{1.3}
\resizebox{\linewidth}{!}{
\begin{tabular}{cc|cccccccccccccc}
\toprule
Baseline                                                                     & Method                 & Spl & RKid & LKid & Gal & Eso & Liv & Sto & Aor & IVC & Vein & Pan & RAG & \multicolumn{1}{l|}{LAG} & Avg.$\uparrow$ \\ \hline
\multirow{9}{*}{\begin{tabular}[c]{@{}c@{}}U-Net~\cite{ronneberger2015u}\\  (ResNet-50)\end{tabular}} & Random~\cite{ronneberger2015u}                 & 91.07 & 91.66 & 91.73 & 41.28 & 71.76 & 95.07 & 73.08 & 87.36 & 79.54 & 65.87 & 66.34 & 64.44 & \multicolumn{1}{l|}{54.04} & 74.86 \\ \cline{2-16} 
                                                                             & ImageNet~\cite{deng2009imagenet}               & 92.87 & 91.16 & 91.87 & 60.00 & 69.83 & 96.19 & 71.08 & 89.80 & 81.84 & 67.59 & 73.99 & 58.03 & \multicolumn{1}{l|}{53.20} & 76.73 \\ \cline{2-16}
                                    
                                      & Inpainting+Contrast+Rotation~\cite{tang2022self}               & 91.62 & 89.85 & 89.30 & 48.94 & 71.81 & 95.35 & 71.56 & 90.32 & 82.70 & 68.49 & 70.77 & \textbf{65.46} & \multicolumn{1}{l|}{57.25} & 76.42 \\ \cline{2-16}
                                                                             & CLIP~\cite{radford2021learning}                   & \textbf{95.01} & 93.20 & 93.14 & 50.91 & 71.11 & \textbf{96.48} & 76.04 & 89.08 & 82.47 & \textbf{71.09} & 65.54 & 64.38 & \multicolumn{1}{l|}{46.70} & 76.55 \\
                                                                                 & CLIP+DenseCLIP~\cite{rao2022denseclip}     & 91.88 & 89.26 & 90.97 & 63.10 & \textbf{73.54} & 95.97 & 80.40 & 90.59 & 82.96 & 68.53 & 72.27 & 61.33 & \multicolumn{1}{l|}{50.02} & 77.76 \\ \cline{2-16} 
                                                                             & PubMedCLIP~\cite{eslami2021does}             & 89.71 & 91.43 & 91.64 & 47.04 & 68.18 & 95.03 & 78.64 & 89.07 & 78.00 & 65.72 & 66.29 & 65.15 & \multicolumn{1}{l|}{57.29} & 75.63 \\
                                                                             & PubMedCLIP+DenseCLIP~\cite{rao2022denseclip} & 93.09 & 90.04 & 90.89 & 62.67 & 69.94 & 95.48 & 81.25 & \textbf{90.77} & 79.95 & 67.27 & 69.35 & 63.01 & \multicolumn{1}{l|}{50.47} & 77.24 \\ \cline{2-16}
                                                                             & \cellcolor[HTML]{C0C0C0}MOSMOS~(Ours) & \cellcolor[HTML]{C0C0C0}94.36 & \cellcolor[HTML]{C0C0C0}\textbf{93.56} & \cellcolor[HTML]{C0C0C0}\textbf{93.81} & \cellcolor[HTML]{C0C0C0}\textbf{73.01} & \cellcolor[HTML]{C0C0C0}72.24 & \cellcolor[HTML]{C0C0C0}96.11 & \cellcolor[HTML]{C0C0C0}\textbf{81.85} & \cellcolor[HTML]{C0C0C0}89.04 & \cellcolor[HTML]{C0C0C0}\textbf{83.72} & \cellcolor[HTML]{C0C0C0}60.74 & \cellcolor[HTML]{C0C0C0}\textbf{76.16} & \cellcolor[HTML]{C0C0C0}64.95 & \multicolumn{1}{l|}{\cellcolor[HTML]{C0C0C0}\textbf{61.68}} & \cellcolor[HTML]{C0C0C0}\textbf{80.10}
                                                                             \\ \cline{2-16}
                                                                        & P-value & \multicolumn{14}{c}{2.7e-2~(Dice)}     \\ \hline
\multirow{9}{*}{\begin{tabular}[c]{@{}c@{}}UNETR~\cite{hatamizadeh2022unetr}\\ (ViT-B/16)\end{tabular}}     & Random~\cite{hatamizadeh2022unetr}                 & \textbf{93.67} & 93.56 & 93.46 & 66.69 & 70.28 & 96.31 & 77.87 & 88.18 & 82.09 & 65.05 & 67.94 & 65.99 & \multicolumn{1}{l|}{59.83} & 78.53 \\ \cline{2-16} 
                                                                             & ImageNet~\cite{deng2009imagenet}               & 91.39 & 93.78 & \textbf{94.00} & 64.13 & 70.52 & 96.48 & 77.82 & \textbf{89.77} & 82.28 & 68.85 & 73.65 & 61.77 & \multicolumn{1}{l|}{\textbf{64.91}} & 79.18 \\ \cline{2-16} 
                                                                             
                                      & Inpainting+Contrast+Rotation~\cite{tang2022self}               & 93.17 & 93.16 & 92.44 & 61.97 & 72.16 & 95.63 & 75.38 & 87.28 & 80.87 & 65.16 & 67.90 & 67.13 & \multicolumn{1}{l|}{61.62} & 77.99 \\ \cline{2-16}         
                                                                             & CLIP~\cite{radford2021learning}                   & 93.08 & \textbf{93.79} & 93.71 & 62.95 & 70.85 & 96.48 & 78.21 & 88.16 & \textbf{83.01} & 68.30 & 73.66 & 66.58 & \multicolumn{1}{l|}{61.58} & 79.26 \\
                                                                             & CLIP+DenseCLIP~\cite{rao2022denseclip}     & 87.97 & 92.25 & 93.00 & 70.64 & 72.58 & 96.31 & 76.13 & 88.43 & 82.38 & 69.88 & 70.20 & \textbf{67.14} & \multicolumn{1}{l|}{60.76} & 79.05 \\ \cline{2-16} 
                                                                             & PubMedCLIP~\cite{eslami2021does}             & 92.75 & 93.27 & 93.08 & 64.96 & 70.84 & 96.10 & 78.27 & 87.97 & 82.18 & 68.18 & 73.55 & 65.57 & \multicolumn{1}{l|}{60.46} & 79.01 \\
                                                                             & PubMedCLIP+DenseCLIP~\cite{rao2022denseclip} & 93.19 & 93.74 & 93.28 & 61.32 & 71.39 & \textbf{96.50} & 79.28 & 88.97 & 82.61 & 69.56 & 74.41 & 66.52 & \multicolumn{1}{l|}{58.97} & 79.21 \\ \cline{2-16}
                                                                             & \cellcolor[HTML]{C0C0C0}MOSMOS~(Ours)  & \cellcolor[HTML]{C0C0C0}93.21 & \cellcolor[HTML]{C0C0C0}93.68 & \cellcolor[HTML]{C0C0C0}93.39 & \cellcolor[HTML]{C0C0C0}\textbf{72.71} & \cellcolor[HTML]{C0C0C0}\textbf{72.96} & \cellcolor[HTML]{C0C0C0}96.46 & \cellcolor[HTML]{C0C0C0}\textbf{80.46} & \cellcolor[HTML]{C0C0C0}88.84 & \cellcolor[HTML]{C0C0C0}82.83 & \cellcolor[HTML]{C0C0C0}\textbf{70.39} & \cellcolor[HTML]{C0C0C0}\textbf{75.00} & \cellcolor[HTML]{C0C0C0}66.90 & \multicolumn{1}{l|}{\cellcolor[HTML]{C0C0C0}57.87} & \cellcolor[HTML]{C0C0C0}\textbf{80.36} 
                                                    \\ \cline{2-16}
                                                                        & P-value & \multicolumn{14}{c}{8.0e-2~(Dice)} \\ \bottomrule
\end{tabular}}
\end{center}
\vspace{-1.5em}
\end{table*}

In the pre-training stage, we resize all 2D images to $H_{1} \times W_{1}=224 \times 224$ as the input resolution and set the token lengths of the medical reports, learnable textual context, and tags to $N=77$, $N_{1}=16$, and $N_{2}=10$, respectively. The feature dimensions of input images, input texts, and outputs are $C_{1}=768$, $C_{2}=512$, and $C=512$, respectively. The network is trained for 50 epochs with a fixed batch size $B_{1}$ of 64, and the optimizer is Adam~\cite{kingma2014adam} with the learning rate of 10$^{-5}$. We compute the validation loss after every epoch and save the checkpoint with the lowest validation loss. 

During the fine-tuning stage, all images are preprocessed following the procedures in~\cite{hatamizadeh2022unetr}. For training, we randomly crop 3D images into a resolution of $H_{2} \times W_{2} \times D_{2} = 96 \times 96 \times 96$. For 2D images, the $D_{2}$ is omitted. We train the whole network using AdamW optimizer~\cite{loshchilov2017decoupled} and set the initial learning rate of 10$^{-4}$ for 5,000 epochs. After 50 epochs, the learning rate is decayed according to the cosine attenuation approach~\cite{he2019bag}. Given the memory constraints, we set the batch size $B_{2}$ to 96 for ResNet-50-based and 2 for ViT-B/16-based methods. The text encoder is fixed to retain more medical language supervision learned from the large-scale image-report pre-training. For inference, we apply the sliding window method with an overlap ratio of 0.5 and keep the same resolution as the training sets. We calculate the evaluation metrics every 100 epochs and select the models with the best values to perform the test.

\subsection{Evaluation metrics}
To objectively evaluate the segmentation performance, we apply the Dice similarity coefficient and Hausdorff Distance 95$\%$~(HD95) as the evaluation metrics. For a given organ category, let $X_v$ and $Y_v$ represent the ground truth and prediction for pixel $v$, and $X^{\prime}$ and $Y^{\prime}$ denote the ground truth and predicted surface point sets. The Dice and HD metrics are defined as:

\vspace{-1.0em}
\begin{equation}
Dice=\frac{2 \sum_{v=1}^V X_v Y_v}{\sum_{v=1}^V X_v+\sum_{v=1}^V Y_v},
\end{equation}

\vspace{-1.0em}
\begin{equation}
\begin{aligned}
HD=\max \left\{\max _{x^{\prime} \in X^{\prime}} \min _{y^{\prime} \in Y^{\prime}} \|x^{\prime}-y^{\prime} \|, \right.
\\
\left. \max _{y^{\prime} \in Y^{\prime}} \min _{x^{\prime} \in X^{\prime}} \|y^{\prime}-x^{\prime} \|\right\},
\end{aligned}
\end{equation}
where Dice measures the overlaps of ground truth and predicted values of $V$ pixels, and HD95 calculates the 95$^{\text{th}}$ percentile of the surface distances between ground truth and predicted point sets.    

\subsection{Quantitative segmentation results}

\begin{figure}[htbp]
\centering
\begin{subfigure}[b]{\columnwidth}
    \includegraphics[width=\columnwidth]{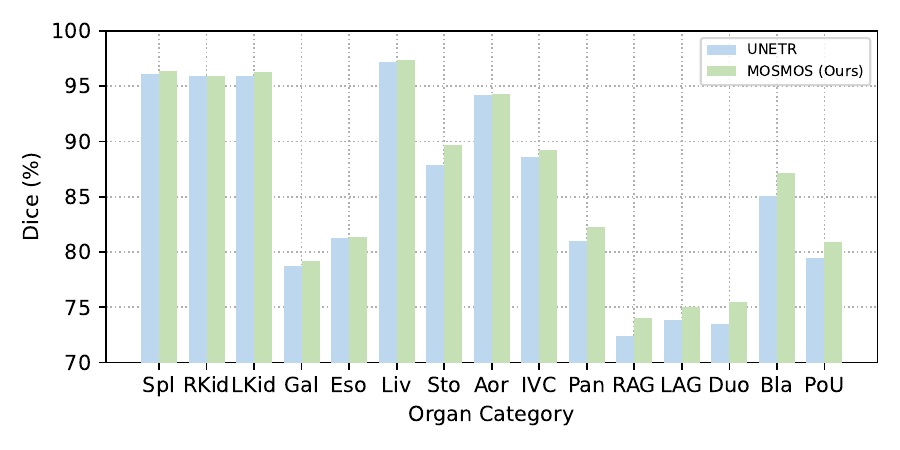}
    \vspace{-1.6em}
    \captionsetup{font=scriptsize}
    \caption{CT}
    \label{ct}
\end{subfigure}
\begin{subfigure}[b]{\columnwidth}
    \includegraphics[width=\columnwidth]{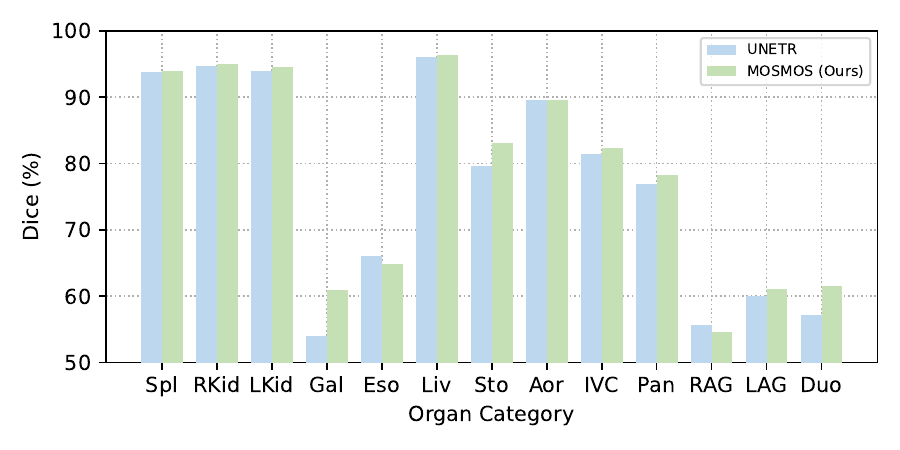}
    \vspace{-1.6em}
    \captionsetup{font=scriptsize}
    \caption{MRI}
    \label{mri}
\end{subfigure}
\caption{The indication of Dice gap between UNETR~(Blue) and MOSMOS~(Green) on AMOS validation sets for CT~(a) and MRI~(b). Notably, Duo, Bla, and PoU belong to open-set organ categories. Note: Spl: spleen, RKid: right kidney, LKid: left kidney, Gal: gallbladder, Eso: esophagus, Liv: liver, Sto: stomach, Aor: aorta, IVC: inferior vena cava, Pan: pancreas, RAG: right adrenal gland, LAG: left adrenal gland, Duo: duodenum, Bla: bladder, PoU: prostate or uterus.}
\label{amos}
\vspace{-2.0em}
\end{figure}

\begin{table*}[htbp]
\caption{Quantitative comparisons of segmentation performance using Dice~($\%$) and HD95~(mm) metrics on MMWHS test set. The best results are bolded. The results of our approach are marked with a gray background color. The p-values are computed based on the average performance of our MOSMOS and PubMedCLIP+DenseCLIP~\cite{rao2022denseclip} in both Dice and HD95 metrics. Note: Myo: myocardium, LA: left atrium, LV: left ventricle, RA: right atrium, RV: right ventricle, AA: ascending aorta, PA: pulmonary artery.}
\vspace{-1.0em}
\label{tab2}
\begin{center}
\renewcommand\arraystretch{1.3}
\resizebox{\linewidth}{!}{
\begin{tabular}{cc|cccccccccccccccc}
\toprule
\multirow{2.5}{*}{Baseline}                                                      & \multirow{2.5}{*}{Method} & \multicolumn{2}{c}{Myo} & \multicolumn{2}{c}{LA} & \multicolumn{2}{c}{LV} & \multicolumn{2}{c}{RA} & \multicolumn{2}{c}{RV} & \multicolumn{2}{c}{AA} & \multicolumn{2}{c|}{PA} & \multicolumn{2}{c}{Avg.} \\ \cmidrule(r){3-4} \cmidrule(r){5-6}  \cmidrule(r){7-8}  \cmidrule(r){9-10}  \cmidrule(r){11-12}  \cmidrule(r){13-14}  \cmidrule{15-16}  \cmidrule{17-18} 
                                                                               &                         & Dice$\uparrow$       & HD95$\downarrow$      & Dice$\uparrow$       & HD95$\downarrow$      & Dice$\uparrow$       & HD95$\downarrow$      & Dice$\uparrow$       & HD95$\downarrow$      & Dice$\uparrow$       & HD95$\downarrow$       & Dice$\uparrow$       & HD95$\downarrow$      & Dice$\uparrow$       & \multicolumn{1}{l|}{HD95$\downarrow$}      & Dice$\uparrow$        & HD95$\downarrow$       \\ \hline
\multirow{9}{*}{\begin{tabular}[c]{@{}c@{}}U-Net~\cite{ronneberger2015u}\\ (ResNet-50)\end{tabular}} & Random~\cite{ronneberger2015u} & 79.92 & 2.89 & 85.00 & 3.47 & 90.57 & 2.99 & 83.35 & 7.16 & 82.92 & 4.74 & 74.43 & \textbf{11.48} & \textbf{71.29} & \multicolumn{1}{l|}{6.12} & 81.07 & 5.55 \\ \cline{2-18} 
                                                                               & ImageNet~\cite{deng2009imagenet}                & 82.76 & 2.46 & \textbf{87.19} & \textbf{3.05} & 93.39 & 2.39 & 85.91 & 6.51 & 90.02 & \textbf{3.16} & 72.88 & 12.04 & 71.10 & \multicolumn{1}{l|}{6.03} & 83.32 & 5.09 \\ \cline{2-18}
                                                                               & Inpainting+Contrast+Rotation~\cite{tang2022self}               & 82.91 & 2.08 & 85.47 & 3.61 & 93.69 & 1.93 & 86.15 & 9.18 & 89.24 & 3.77 & 73.74 & 11.84 & 64.37 & \multicolumn{1}{l|}{8.33} & 82.22 & 5.82 \\ \cline{2-18}  
                                                                               & CLIP~\cite{radford2021learning}                    & 81.87 & 2.49 & 84.66 & 3.46 & 92.79 & 3.23 & 81.17 & 7.65 & 82.64 & 5.72 & 71.18 & 11.55 & 70.31 & \multicolumn{1}{l|}{\textbf{4.90}} & 80.66 & 5.57 \\
                                                                               & CLIP+DenseCLIP~\cite{rao2022denseclip} & 82.30 & 2.46 & 86.34 & 3.18 & 93.74 & 1.78 & 78.77 & \textbf{5.76} & 87.44 & 3.45 & 73.02 & 12.22 & 71.07 & \multicolumn{1}{l|}{5.15} & 81.81 & \textbf{4.86} \\ \cline{2-18} 
                                                                               & PubMedCLIP~\cite{eslami2021does}              & 79.74 & 2.55 & 81.55 & 4.06 & 93.50 & 2.19 & 81.00 & 7.99 & 87.05 & 5.15 & 71.09 & 13.13 & 65.26 & \multicolumn{1}{l|}{6.90} & 79.88 & 5.99 \\
                                                                               & PubMedCLIP+DenseCLIP~\cite{rao2022denseclip} & 81.39 & 2.35 & 84.95 & 3.24 & 93.48 & 2.16 & 85.80 & 9.32 & 88.65 & 4.01 & 69.13 & 15.59 & 65.10 & \multicolumn{1}{l|}{10.38} & 81.21 & 6.72 \\ \cline{2-18}
                                                                               & \cellcolor[HTML]{C0C0C0}\textbf{MOSMOS~(Ours)} & \cellcolor[HTML]{C0C0C0}\textbf{83.74} & \cellcolor[HTML]{C0C0C0}\textbf{2.00} & \cellcolor[HTML]{C0C0C0}84.99 & \cellcolor[HTML]{C0C0C0}3.46 & \cellcolor[HTML]{C0C0C0}\textbf{93.78} & \cellcolor[HTML]{C0C0C0}\textbf{1.70} & \cellcolor[HTML]{C0C0C0}\textbf{86.51} & \cellcolor[HTML]{C0C0C0}6.76 & \cellcolor[HTML]{C0C0C0}\textbf{90.89} & \cellcolor[HTML]{C0C0C0}3.25 & \cellcolor[HTML]{C0C0C0}\textbf{74.87} & \cellcolor[HTML]{C0C0C0}13.22 & \cellcolor[HTML]{C0C0C0}70.24 & \multicolumn{1}{l|}{\cellcolor[HTML]{C0C0C0}6.81} & \cellcolor[HTML]{C0C0C0}\textbf{83.57} & \cellcolor[HTML]{C0C0C0}5.31\\ \cline{2-18}
                                                                        & P-values & \multicolumn{16}{c}{1.6e-2~(Dice), 3.1e-2~(HD95)} 
                                                    \\ \hline
\multirow{9}{*}{\begin{tabular}[c]{@{}c@{}}UNETR~\cite{hatamizadeh2022unetr}\\ (ViT-B/16)\end{tabular}}       & Random~\cite{hatamizadeh2022unetr} & 84.21 & 1.83 & 83.96 & 4.02 & 93.85 & 1.93 & 86.20 & 7.08 & 91.72 & 2.88 & 82.27 & 6.01 & 76.46 & \multicolumn{1}{l|}{8.25} & 85.52 & 4.57 \\ \cline{2-18} 
                                                                               & ImageNet~\cite{deng2009imagenet}                & 83.96 & 1.91 & 83.38 & 4.58 & 93.35 & 2.23 & 84.39 & 6.84 & 89.96 & 3.56 & 82.82 & 8.19 & 77.62 & \multicolumn{1}{l|}{7.61} & 85.07 & 4.99 \\ \cline{2-18}
                                                                               & Inpainting+Contrast+Rotation~\cite{tang2022self}               & \textbf{85.03} & \textbf{1.77} & 87.16 & 3.47 & \textbf{94.06} & \textbf{1.83} & 87.18 & 6.31 & 92.42 & 2.48 & 82.91 & 5.22 & 77.87 & \multicolumn{1}{l|}{5.69} & 86.66 & 3.82 \\ \cline{2-18}
                                                                               & CLIP~\cite{radford2021learning}                    & 84.72 & 1.83 & 86.88 & 3.55 & 94.03 & 1.93 & 87.19 & 4.85 & 92.31 & 2.63 & 83.54 & 5.37 & 78.74 & \multicolumn{1}{l|}{6.86} & 86.77 & 3.86 \\
                                                                               & CLIP+DenseCLIP~\cite{rao2022denseclip} & 84.74 & \textbf{1.77} & 86.54 & \textbf{3.44} & 93.83 & \textbf{1.83} & 87.85 & 5.35 & 92.01 & 3.11 & \textbf{84.80} & 5.08 & 78.35 & \multicolumn{1}{l|}{5.47} & 86.87 & 3.72 \\ \cline{2-18}                      
                                                                               & PubMedCLIP~\cite{eslami2021does}              & 84.67 & 1.83 & 85.65 & 4.00 & 93.91 & 1.85 & 87.58 & \textbf{4.36} & 92.93 & 2.76 & 83.53 & 5.75 & 78.84 & \multicolumn{1}{l|}{\textbf{5.33}} & 86.73 & 3.70 \\
                                                                               & PubMedCLIP+DenseCLIP~\cite{rao2022denseclip} & 82.69 & 2.24 & 79.35 & 5.11 & 92.86 & 2.58 & 82.60 & 8.05 & 88.75 & 10.47 & 79.88 & 7.69 & 69.36 & \multicolumn{1}{l|}{8.90} & 82.21 & 6.44 \\ \cline{2-18}
                                                                               & \cellcolor[HTML]{C0C0C0}\textbf{MOSMOS~(Ours)} & \cellcolor[HTML]{C0C0C0}84.65 & \cellcolor[HTML]{C0C0C0}\textbf{1.77} & \cellcolor[HTML]{C0C0C0}\textbf{87.62} & \cellcolor[HTML]{C0C0C0}3.48 & \cellcolor[HTML]{C0C0C0}94.02 & \cellcolor[HTML]{C0C0C0}1.85 & \cellcolor[HTML]{C0C0C0}\textbf{88.61} & \cellcolor[HTML]{C0C0C0}4.62 & \cellcolor[HTML]{C0C0C0}\textbf{93.22} & \cellcolor[HTML]{C0C0C0}\textbf{2.30} & \cellcolor[HTML]{C0C0C0}84.08 & \cellcolor[HTML]{C0C0C0}\textbf{4.60} & \cellcolor[HTML]{C0C0C0}\textbf{79.49} & \multicolumn{1}{l|}{\cellcolor[HTML]{C0C0C0}5.60} & \cellcolor[HTML]{C0C0C0}\textbf{87.38} & \cellcolor[HTML]{C0C0C0}\textbf{3.46}
                                                    \\ \cline{2-18}
                                                                        & P-values & \multicolumn{16}{c}{1.6e-2~(Dice), 1.6e-2~(HD95)} \\ \bottomrule
\end{tabular}}
\end{center}
\vspace{-1.5em}
\end{table*}

\subsubsection{Abdominal multi-organ segmentation on BTCV}
As shown in Table \ref{tab1}, we report the abdominal multi-organ segmentation results of our MOSMOS and other approaches with two different baselines on BTCV. We see that the pre-training methods generally perform better than training from scratch. Compared with other pre-training methods, MOSMOS consistently attains the highest Dice scores, both on average and across the majority of organ categories. This noteworthy accomplishment is attributed to the incorporation of two key components during the pre-training phase: global image-report alignment and local pixel-tag alignment. Specifically, our MOSMOS is 3.37$\%$ and 1.18$\%$ Dice higher than the ImageNet-based pre-training~\cite{deng2009imagenet} on ResNet-50 and ViT-B/16 visual backbones, respectively. MOSMOS also surpasses the state-of-the-art self-supervised pre-trained baselines~(denoted by Inpainting+Contrast+Rotation~\cite{tang2022self}) by 3.68$\%$ and 2.37$\%$ on average of 13 organs. Besides, MOSMOS consistently maintains advantages of at least 1.10$\%$ with respect to these contrastive language-image pre-training models~\cite{radford2021learning, rao2022denseclip, eslami2021does}. Although MOSMOS with ViT-B/16 visual backbone does not improve as much as with ResNet-50, it outperforms using ResNet-50, so ViT is more suitable for multi-organ segmentation tasks. As for why MOSMOS does not perform best in some organs, we consider that the feature extraction differences in visual backbones affect the positioning capability of attention maps. 

\subsubsection{Abdominal multi-organ segmentation on AMOS}
A performance comparison of multi-organ segmentation tasks on the AMOS dataset for both CT and MRI modalities using MOSMOS versus the baseline UNETR~\cite{hatamizadeh2022unetr} is presented in Fig. \ref{amos}. As depicted in Fig. \ref{ct}, MOSMOS consistently outperforms UNETR across all CT segmentation tasks on AMOS, with an average Dice score improvement from 85.37$\%$ to 86.29$\%$. Significant improvements can be observed in the closed-set tasks of the stomach, pancreas, right adrenal gland, left adrenal gland, and the open-set tasks of the duodenum, bladder, prostate or uterus, with Dice scores advancing from 87.89$\%$ to 89.63$\%$, 80.99$\%$ to 82.24$\%$, 72.38$\%$ to 74.07$\%$, 73.88$\%$ to 74.98$\%$, 73.47$\%$ to 75.50$\%$, 85.06$\%$ to 87.11$\%$, and 79.48$\%$ to 80.87$\%$, respectively. In Fig. \ref{mri}, for all MRI tasks on AMOS, the average Dice score increases from 76.84$\%$ to 78.17$\%$. Distinct improvements are evident in the closed-set stomach category and the open-set duodenum category, with Dice scores improving from 79.58$\%$ to 83.05$\%$ and 57.19$\%$ to 61.48$\%$, respectively. The gallbladder category in the closed-set displays the most substantial improvement, with a Dice score of 60.99$\%$ compared to 54.08$\%$.

\subsubsection{Cardiac substructure segmentation on MMWHS}
Table \ref{tab2} presents the class-specific results in both Dice and HD95 metrics on cardiac substructure segmentation using the MMWHS dataset. Compared to previous approaches, the proposed MOSMOS displays more strength in Dice than in HD95. Specifically speaking, our MOSMOS outperforms other methods in 9 out of 14 categories in Dice under two different baselines, while 5 out of 14 categories in HD95. From the perspective of average performance, we notice that MOSMOS often achieves better capability. For instance, we gain the state-of-the-art Dice of 83.57$\%$ and 87.38$\%$ when adopting U-Net~\cite{ronneberger2015u} and UNETR~\cite{hatamizadeh2022unetr} as the segmentation baselines, respectively, while keeping the lowest HD95 in UNETR baseline. Moreover, MOSMOS surpasses DenseCLIP~\cite{rao2022denseclip}---the best-performing contrastive language-image pre-training approach---by over 0.51$\%$ in Dice.

\subsubsection{Statistical significance}
In Table \ref{tab1} and Table \ref{tab2}, we employ Wilcoxon signed rank test to calculate p-values between the average performance of our MOSMOS and PubMedCLIP+DenseCLIP~\cite{rao2022denseclip} in both Dice and HD95 metrics. As we can see, MOSMOS demonstrates statistically significant performance, yielding p-values below 5e-2 across both Dice and HD95 metrics on two distinct baselines~(U-Net \& UNETR) and two public datasets~(BTCV \& MMWHS). The sole exception is observed with UNETR baseline on the BTCV dataset. These findings indicate that, in general, MOSMOS has significant advantages over PubMedCLIP+DenseCLIP.

\subsection{Analytical ablation studies}

\begin{table}[ht]
\caption{Exploration of the effects of different modules used in MOSMOS. \textbf{MLR} and \textbf{IRC} denote the multi-label recognition loss and the image-report contrastive loss applied in the pre-training stage, respectively. \textbf{Prompt} represents the learnable textual context. \textbf{CLIP} refers to the introduction of the pre-trained CLIP parameters. We highlight the best results in bold front.}
\vspace{-1.0em}
\label{tab3}
\begin{center}
\resizebox{\linewidth}{!}{
\begin{tabular}{ccccccc}
\toprule
Row & MLR & IRC & Prompt & CLIP & Avg. Dice$\uparrow$ & Avg. HD95$\downarrow$ \\ \midrule
0            & $\checkmark$ & $\checkmark$ & $\checkmark$  & $\checkmark$       &    \textbf{80.36}                          &    \textbf{6.39}                            \\
1            & $\checkmark$ &              & $\checkmark$  & $\checkmark$       &               79.82               &           7.98                     \\
2            & $\checkmark$ & $\checkmark$ &               & $\checkmark$       &              79.06                 &           8.03                   \\
3            & $\checkmark$ & $\checkmark$ & $\checkmark$  &                    &               78.77               &              9.94                  \\
4            &  &  & $\checkmark$  &        &               75.01               &              42.95                   \\ \bottomrule
\end{tabular}}
\end{center}
\vspace{-1.5em}
\end{table}

\subsubsection{Different modules in MOSMOS}
We provide a thorough empirical study on MOSMOS by removing the individual modules in Table \ref{tab3}. For simplicity, we conduct experiments on the BTCV dataset and choose UNETR as the default baseline. 

\begin{figure}[htbp]
\centering
\begin{subfigure}[b]{\columnwidth}
    \includegraphics[width=\columnwidth]{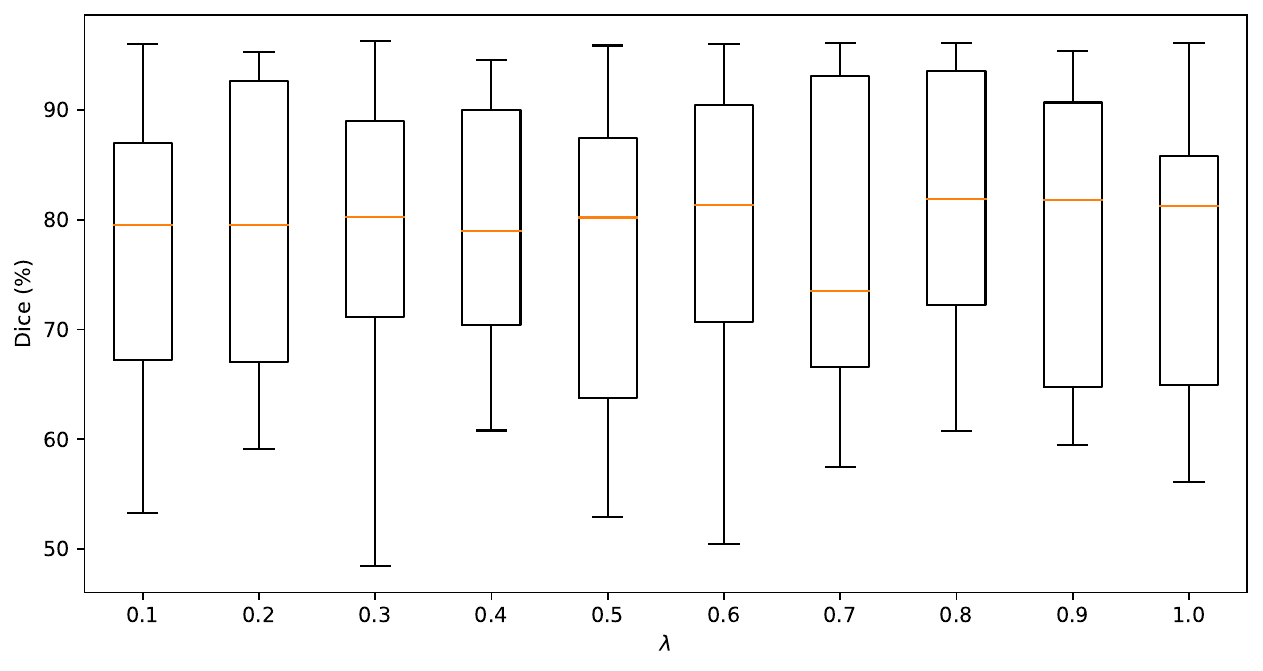}
    \vspace{-1.8em}
    \captionsetup{font=scriptsize}
    \caption{ResNet-50}
    \label{ResNet-50}
\end{subfigure}
\begin{subfigure}[b]{\columnwidth}
    \includegraphics[width=\columnwidth]{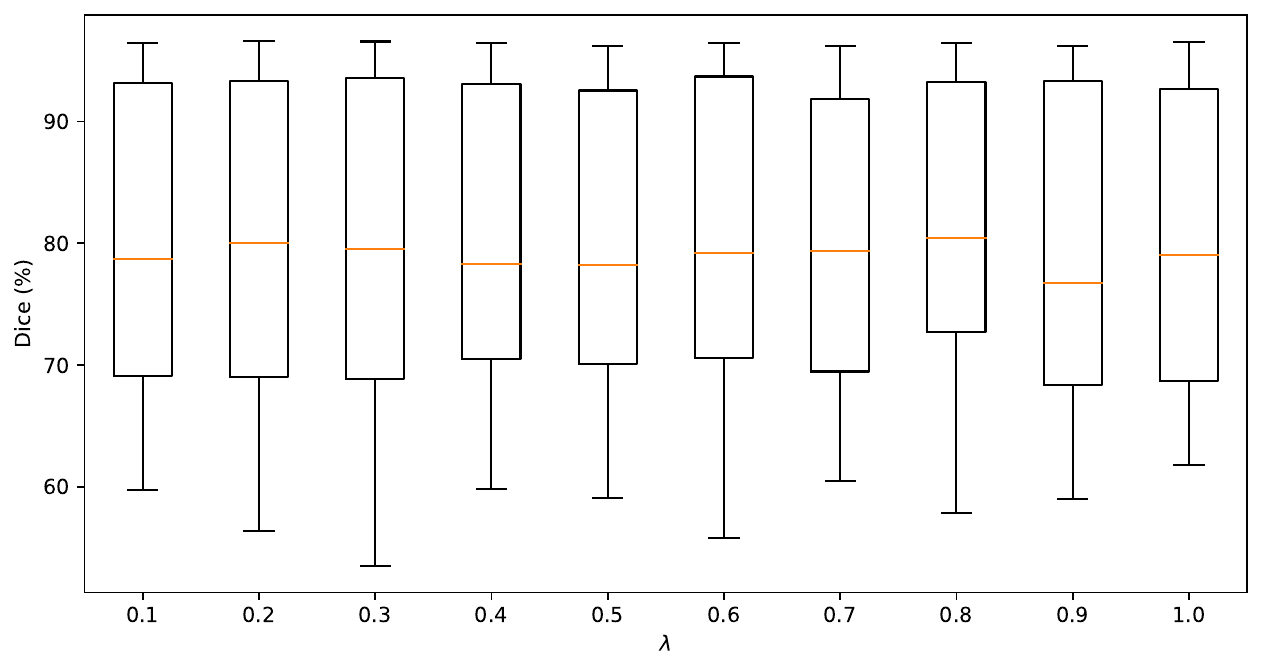}
    \vspace{-1.8em}
    \captionsetup{font=scriptsize}
    \caption{ViT-B/16}
    \label{ViT-B/16}
\end{subfigure}
\caption{Dice box plots of our approach based on ResNet-50~(a) and ViT-B/16~(b) visual backbones for BTCV, when varying the weight $\lambda$ of pixel-tag aligning loss from 0.1 to 1.0.}
\label{box_plots}
\vspace{-1.0em}
\end{figure}

First, we investigate the effect of medical image-report contrastive learning in the pre-training stage. By comparing row 1 with row 0, we observe that dropping the cross-modal contrastive task would adversely affect the overall performance by 0.54$\%$ and 1.59mm in average Dice and HD95, respectively. We argue the reason behind this is that the global image-report aligning is a prerequisite for the local pixel-tag aligning and thus benefits downstream segmentation tasks. Next, we remove the learnable textual context so that each organ tag is embedded alone by the text encoder~(row 2). Such an operation causes Dice to drop by 1.30$\%$ and HD95 to rise by 1.64mm~(compared with row 0). This phenomenon demonstrates the helpfulness of mitigating the gaps between organ tags and reports. In addition, we consider not adopting CLIP parameters for initialization~(row 3), where we can see a 1.59$\%$ decrease in Dice and a 3.55mm increase in HD95. This result verifies the advantage of large-scale cross-modal pre-training. Last but not least, we evaluate the significance of the entire pre-training process. A comparison between Row 4 and Row 0 reveals that the integration of supervision derived from medical reports can markedly improve overall performance. Such an improvement is attributed to the introduction of comprehensive medical prior knowledge without any additional manpower expense.

\begin{table}[ht]
\caption{Segmentation performance comparisons of the ViT-B backbone with different patch resolutions using Dice~($\%$) metric. We highlight the best results in bold front.}
\vspace{-1.5em}
\label{tab4}
\begin{center}
\resizebox{\linewidth}{!}{
\begin{tabular}{ccccc}
\toprule
\multirow{2.5}{*}{\begin{tabular}[c]{@{}c@{}}Patch \\ Resolution\end{tabular}} & \multicolumn{2}{c}{BTCV} & \multicolumn{2}{c}{MMWHS} \\ \cmidrule(r){2-3}  \cmidrule(r){4-5}  
                                  & UNETR~\cite{hatamizadeh2022unetr} & \textbf{MOSMOS} & UNETR~\cite{hatamizadeh2022unetr} & \textbf{MOSMOS} \\ \midrule
32                                & 77.13 & \textbf{78.24} & 77.86 & \textbf{80.79} \\ \midrule
16                                & 78.53 & \textbf{80.36} & 85.52 & \textbf{87.38} \\ \bottomrule
\end{tabular}}
\end{center}
\vspace{-2.0em}
\end{table}

\subsubsection{Different weights of the pixel-tag aligning loss}
We vary the weight of pixel-tag aligning loss to explore the sensitivity of results to the trade-off parameter $\lambda$ in Eq.~(\ref{eq_total}). To be specific, we range $\lambda \in \left[0.1, 1.0\right]$ at a step of 0.1 and analyze the organ-wise segmentation performance on the BTCV dataset. As shown in Fig. \ref{ResNet-50}, the box plot displays the average Dice across each organ of our method based on the ResNet-50 visual backbone. Our MOSMOS achieves the best performance on the BTCV test set when the $\lambda$ is set to 0.8. The performance fluctuations are very small, except when $\lambda$ is 0.7. In comparison, our MOSMOS is capable of generally outperforming the baseline~(e.g., 74.86$\%$ for U-Net shown in Table \ref{tab1}). In Fig. \ref{ViT-B/16}, we compare the performance of MOSMOS based on the ViT-B/16 visual backbone with different $\lambda$. Although MOSMOS attains the highest Dice score when $\lambda$ equals 0.8, it exhibits low sensitivity to $\lambda$. Considering the performance across both visual backbones, we empirically set $\lambda$ to 0.8.

\subsubsection{Different patch resolutions of the ViT-B backbone}
In Table \ref{tab4}, we compare the average performance of the ViT-B visual backbone with different input patch resolutions. It shows that the performance significantly improves when decreasing the patch resolution. Specifically, dropping the resolution from 32 to 16 boosts the Dice of our MOSMOS by 2.12$\%$ and 6.59$\%$ on BTCV and MMWHS datasets, respectively. We can also observe that the proposed MOSMOS consistently maintains a significant advantage over the baseline UNETR in different resolutions and datasets. However, a lower patch resolution leads to a longer sequence and, therefore, higher memory cost. Considering the trade-off between segmentation performance and memory consumption, we empirically set the input patch resolution of ViT-B to 16.  

\subsubsection{Different label ratios in the fine-tuning stage}
\begin{figure}[ht]
\centerline{\includegraphics[width=\columnwidth]{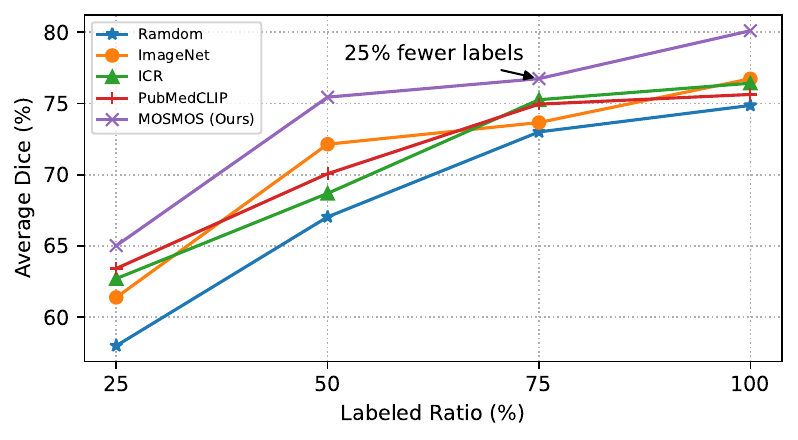}}
\caption{Performance of different approaches under various label ratios on BTCV test dataset. We highlight the percentage of annotated training data in the fine-tuning stage that our MOSMOS needs to achieve results comparable with those obtained by training from scratch or using other pre-training methods. Note that all five methods utilize the same ResNet-50 visual backbone. Note: ICR: Inpainting+Contrast+Rotation.}
\label{label_ratios}
\vspace{-1.5em}
\end{figure}

Fig. \ref{label_ratios} displays the performance comparison of various approaches under different label ratios on BTCV test dataset. Using only 25$\%$ of labeled data, our MOSMOS achieves a 7$\%$ improvement in performance compared to training a model from scratch. When utilizing the full set of labeled data, MOSMOS outperforms models trained from scratch or those using other pre-training methods by an average Dice score increase of 3.37$\%$. Notably, MOSMOS only needs 75$\%$ of  the annotated training data to match the performance comparable with those of other methods under a 100$\%$ labeled ratio. This highlights MOSMOS's efficiency in reducing annotation efforts by approximately 25$\%$ for the multi-organ segmentation task on BTCV.

\subsection{Visualization for qualitative segmentation results}
\begin{figure}[ht]
\centerline{\includegraphics[width=\columnwidth]{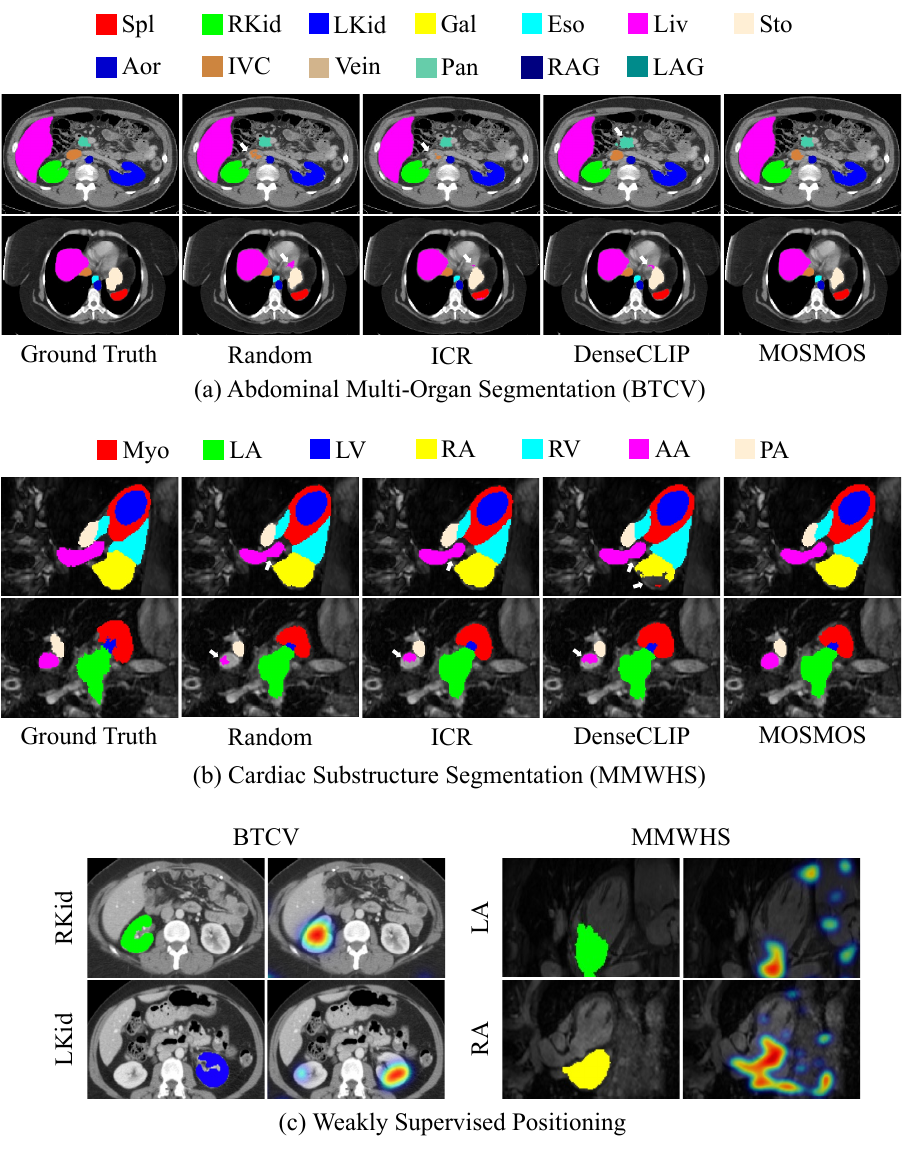}}
\caption{Visualization of segmentation and weakly supervised positioning results. We mainly compare different pre-training strategies for the UNETR baseline on two widely used datasets. Note: ICR: Inpainting+Contrast+Rotation, DenseCLIP: PubMedCLIP+DenseCLIP, Spl: spleen, RKid: right kidney, LKid: left kidney, Gal: gallbladder, Eso: esophagus, Liv: liver, Sto: stomach, Aor: aorta, IVC: inferior vena cava, Vein: portal vein and splenic vein, Pan: pancreas, RAG: right adrenal gland, LAG: left adrenal gland, Myo: myocardium, LA: left atrium, LV: left ventricle, RA: right atrium, RV: right ventricle, AA: ascending aorta, PA: pulmonary artery. Error predictions are pointed by white arrows~(best views in color).}
\label{visualization}
\vspace{-1.8em}
\end{figure}

Fig. \ref{visualization} illustrates the segmentation and weakly supervised positioning results for qualitative evaluation. We mainly visualize the segmentation maps of our MOSMOS and other pre-training approaches on two public datasets. Compared to training from scratch and self-supervised or image-report contrastive pre-training, our MOSMOS displays visual improvements in capturing the shape of inferior vena cava~(IVC, row 1 on BTCV) and pancreas~(Pan, row 1 on BTCV), right atrium~(RA, row 1 on MMWHS), and ascending aorta~(AA, row 1 and row 2 on MMWHS). In addition, MOSMOS can reduce the prediction of false positives. One representative example is the second case on BTCV. Other methods predict the wrong liver~(Liv) pixels near the stomach~(Sto). Furthermore, the weakly supervised positioning results show that MOSMOS can distinguish between left and right organs through tailored pre-training tasks, demonstrating the superiority of our approach. 

\section{Discussion}
\subsection{Strengths}
We verify the effectiveness and generalization of our MOSMOS on multi-disease, multi-modal, and multi-organ datasets. Considering the cost of collection and annotation, BTCV, AMOS, and MMWHS datasets consist of a small number of annotated images, which cannot train effective models using randomly initialized weights from scratch. Thus we make a thorough comparison of the pre-training strategies. We observe that MOSMOS outperforms the previous pre-training approaches on most of the average metrics and organ categories. The main reasons for this are: (i) The ImageNet-based pre-training utilizes nature images, which exit enormous differences from medical images. In addition, this method is supervised and requires extensive annotations. (ii) Without the need for annotation effort, the self-supervised pre-training approach designs proxy tasks to learn solely visual representations, which does not introduce additional potentially exploitable supervisory information and has a gap with downstream tasks. (iii) As for the image-report contrastive pre-training, it adopts language priors paired with medical images as supervision without extra human effort. However, the visual spatial features transferred downstream are indirectly aligned to the text embeddings via the visual global features. In contrast, MOSMOS directly aligns the visual spatial features and the tag embeddings corresponding to the organ tags by introducing multi-label recognition in the pre-training stage, which can roughly identify the same organ with different shapes and sizes using attention maps in the Transformer decoder. Unlike the traditional multi-label classification, which encodes the multiple labels into a string of numbers as input, we take the embeddings of multiple organ tags as input. In this way, our MOSMOS is scalable and generalized. Meanwhile, MOSMOS is suitable for any segmentation model. The performance improvements on U-Net and UNETR demonstrate the universality of the MOSMOS framework. 

\subsection{Limitations}
Our approach still has some limitations that can be improved in future works. 

First, the proposed MOSMOS has only been pre-trained using 2D medical image-report pairs and transferred to 2D and 3D multi-organ segmentation tasks. This is mainly due to the lack of publicly available 3D image-report pairs, which are more consistent with clinical practice in most medical imaging modalities. In future effort, we will extend our framework to 3D image-report pre-training by constructing this dataset. 

Second, in the current pre-training stage, we have constructed only 20 limited organ tag categories, primarily focused on abdominal multi-organs and cardiac substructures, which are not sufficient for fine-grained segmentations of the entire complex human body organs. Despite this, the diversity of medical reports in the pre-training stage provides a preliminary basis, as demonstrated in Fig. \ref{amos}, for MOSMOS to exhibit a degree of open-set segmentation capability for abdominal organs on the AMOS dataset. Therefore, we can further refine our approach by expanding the tag list used in the pre-training stage and developing more advanced algorithms for open-set multi-organ segmentation. 

\begin{table}[ht]\tiny
\caption{Segmentation performance comparisons of UNETR and MOSMOS using Dice~($\%$) metric on BRATS test set. We highlight the best results in bold front. Note: TC: Tumor Core, WT: Whole Tumor, ET: Enhancing Tumor.}
\label{tab_brats}
\vspace{-1.5em}
\begin{center}
\renewcommand\arraystretch{1.25}
\resizebox{\linewidth}{!}{
\begin{tabular}{ccccc}
\hline
Method & TC & WT & ET & Avg.$\uparrow$  \\ \hline
UNETR~\cite{hatamizadeh2022unetr}  & \textbf{66.70}      & 83.31       & \textbf{69.38}           & 73.13 \\
MOSMOS & 65.79      & \textbf{84.43}       & 69.27           & \textbf{73.16} \\ \hline
\end{tabular}}
\end{center}
\vspace{-2.5em}
\end{table}

Third, we mainly focus on organ segmentation, but ignore the descriptions of lesion morphology, size, location, and number in the reports, which can guide more significant tasks of fine-grained lesion segmentation. To further explore the generalization of our MOSMOS model, we extend its application to a slightly out-of-domain task---brain tumor segmentation on the BRATS dataset. Table \ref{tab_brats} shows the performance comparison in Dice score between MOSMOS and the baseline UNETR. MOSMOS surpasses UNETR by 1.12$\%$ Dice in whole tumor segmentation, yet exhibits comparable or inferior performance in the more granular tumor core and enhancing tumor segmentation tasks. Due to the significant differences between organs and lesions, the performance improvement in open-set brain tumor segmentation on BRATS is not substantial. Consequently, we aim to optimize our framework to be more suitable for lesion segmentation by mining the medical reports for more detailed information to further demonstrate its generality.

\section{Conclusions}
In this paper, we present a novel framework, dubbed MOSMOS, for multi-organ segmentation by leveraging cross-modal pre-training with medical image-report pairs. Based on global image-report aligning, MOSMOS first introduces the proxy task of local pixel-tag aligning. It utilizes a multi-label recognition approach to position the organ tags extracted from reports in the corresponding images, which is more suitable for complex fine-grained segmentation tasks in the downstream. Thus, the proposed framework is capable of being general for multi-disease, multi-modal, and multi-organ segmentation tasks on both 2D and 3D networks.

% Numbered list
% Use the style of numbering in square brackets.
% If nothing is used, default style will be taken.
% \begin{enumerate}[a)]
% \item 
% \item 
% \item 
% \end{enumerate}  

% Unnumbered list
% \begin{itemize}
% \item 
% \item 
% \item 
% \end{itemize}  

% Description list
% \begin{description}
% \item[]
% \item[] 
% \item[] 
% \end{description}  

% Figure
% \begin{figure}[<options>]
% 	\centering
% 		\includegraphics[<options>]{}
% 	  \caption{}\label{fig1}
% \end{figure}

% \begin{table}[<options>]
% \caption{}\label{tbl1}
% \begin{tabular*}{\tblwidth}{@{}LL@{}}
% \toprule
%   &  \\ % Table header row
% \midrule
%  & \\
%  & \\
%  & \\
%  & \\
% \bottomrule
% \end{tabular*}
% \end{table}

% Uncomment and use as the case may be
%\begin{theorem} 
%\end{theorem}

% Uncomment and use as the case may be
%\begin{lemma} 
%\end{lemma}

%% The Appendices part is started with the command \appendix;
%% appendix sections are then done as normal sections
\appendix

\section*{CRediT authorship contribution statement}
\textbf{Weiwei Tian:} Writing – review \& editing, Writing – original draft, Visualization, Validation, Supervision, Software, Resources, Project administration, Methodology, Investigation, Formal analysis, Data curation, Conceptualization. \textbf{Xinyu Huang:} Writing – review \& editing, Writing – original draft, Visualization, Validation, Supervision, Software, Resources, Methodology, Investigation, Data curation, Conceptualization. \textbf{Junlin Hou:} Writing – review \& editing, Writing – original draft, Visualization, Validation, Software, Methodology, Formal analysis. \textbf{Caiyue Ren:} Writing – review \& editing, Writing – original draft, Visualization, Validation, Software, Formal analysis. \textbf{Longquan Jiang:} Writing – review \& editing, Writing – original draft, Visualization, Validation, Supervision, Software, Resources, Project administration, Investigation, Funding acquisition. \textbf{Rui-Wei Zhao:} Writing – review \& editing, Writing – original draft, Visualization, Validation, Funding acquisition, Formal analysis. \textbf{Gang Jin:} Writing – review \& editing, Writing – original draft, Supervision, Conceptualization. \textbf{Yuejie Zhang:} Writing – review \& editing, Writing – original draft, Supervision, Funding acquisition, Formal analysis. \textbf{Daoying Geng:} Writing – review \& editing, Writing – original draft, Supervision, Investigation, Formal analysis.

\section*{Declaration of competing interest}
The authors declare that they have no competing interests.

\section*{Data availability}
Data will be made available on request.

\section*{Acknowledgments}
This work was supported in part by the Science and Technology Commission of Shanghai Municipality~(No.22511106003, No.23511100602) and the Shanghai Research and Innovation Functional Program under Grant 17DZ2260900.

% \section{}\label{}

% To print the credit authorship contribution details
% \printcredits

%% Loading bibliography style file
% \bibliographystyle{model1-num-names}
% \bibliographystyle{cas-model2-num-names}
\bibliographystyle{cas-model2-names}

% Loading bibliography database
\bibliography{cas-refs}

% Biography
% \bio{}
% % Here goes the biography details.
% \endbio

% \bio{pic1}
% % Here goes the biography details.
% \endbio

\end{sloppypar}
\end{document}